%% file: main.tex
\pdfoutput=1
\documentclass[10pt,twocolumn,letterpaper]{article}

\PassOptionsToPackage{table}{xcolor}
\usepackage[pagenumbers]{wacv}
\usepackage{pifont}
\usepackage{makecell}
\usepackage{multirow}
\usepackage{enumitem}
\usepackage{dblfloatfix}
\usepackage{float}
\usepackage{algorithm}
\usepackage{algorithmic}

\usepackage{listings}
\renewcommand{\dbltopfraction}{0.9}
\renewcommand{\dblfloatpagefraction}{0.7}
\renewcommand{\baselinestretch}{0.918}

\makeatletter
\renewcommand\paragraph{\@startsection{paragraph}{4}{\z@}%
  {\dimexpr 3.25ex-0.1cm\relax \@plus1ex \@minus.2ex}%
  {-1em}%
  {\normalfont\normalsize\bfseries}}
\makeatother

\def\wacvPaperID{238}
\def\confName{WACV}
\def\confYear{2027}

\definecolor{wacvblue}{rgb}{0.21,0.49,0.74}
\usepackage[pagebackref,breaklinks,colorlinks,allcolors=wacvblue]{hyperref}
\hypersetup{
  pdftitle={HiMu: Hierarchical Multimodal Frame Selection for Long Video Question Answering},
  pdfauthor={Dan Ben-Ami, Gabriele Serussi, Kobi Cohen, Chaim Baskin}
}

\newcommand{\cmark}{{\color{green!70!black}\ding{51}}}
\newcommand{\xmark}{{\color{red}\ding{55}}}

\definecolor{revblue}{rgb}{0.0,0.0,0.0}
\newcommand{\revised}[1]{\texorpdfstring{\textcolor{revblue}{#1}}{#1}}
\newcommand{\revon}{\color{revblue}}
\newcommand{\revoff}{\color{black}}
\renewcommand{\revised}[1]{#1}
\renewcommand{\revon}{}
\renewcommand{\revoff}{}

\makeatletter
\newcommand{\himuTeaserFigure}{%
  \begin{center}
    \includegraphics[width=\textwidth]{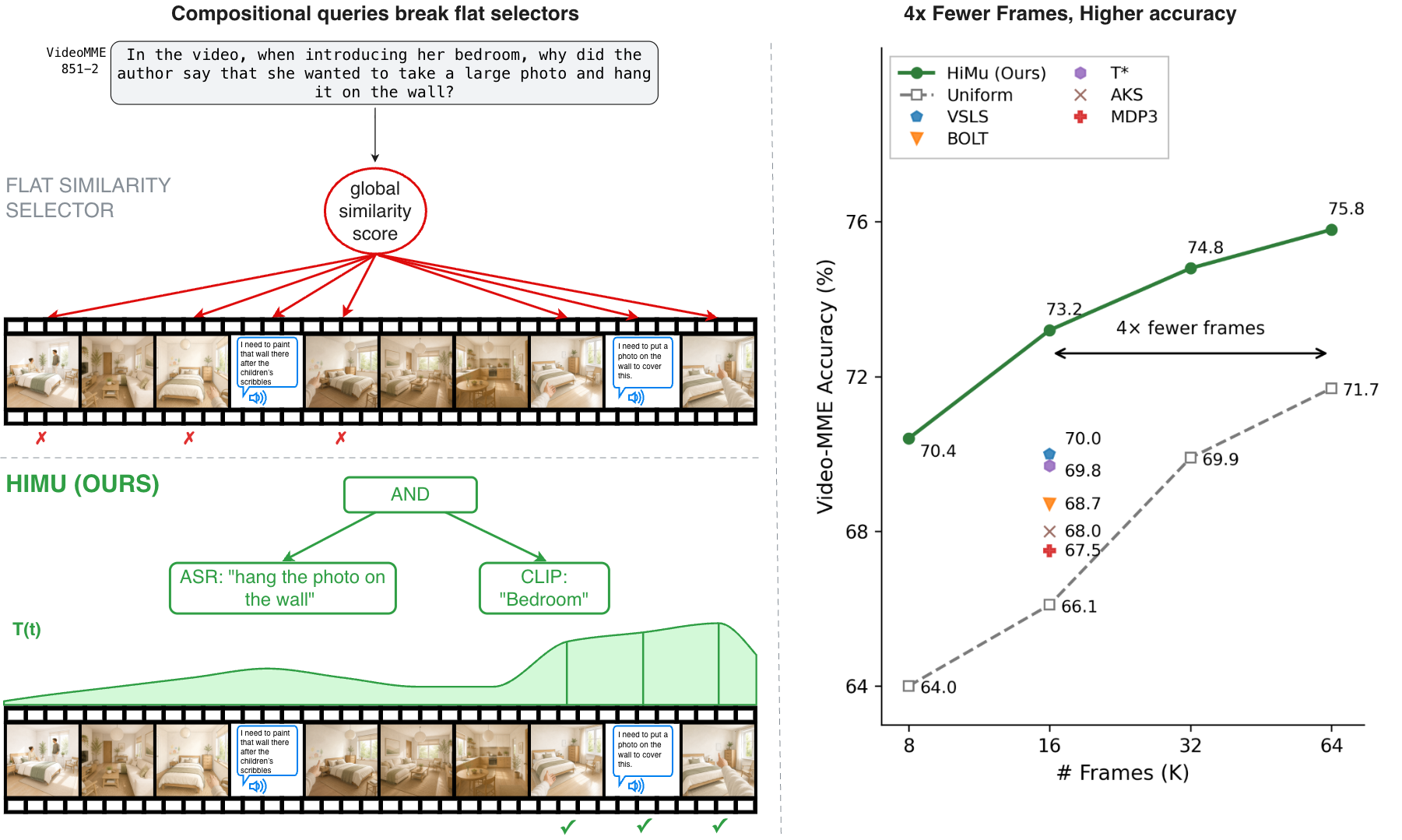}%
  \end{center}%
  \refstepcounter{figure}%
  \label{fig:himu-teaser}%
  \vspace{-0.7cm}%
  \@makecaption{\figurename~\thefigure}{\textbf{HiMu: compositional, multimodal frame selection under a controlled frame budget.}
  \textbf{Left:} flat similarity selectors collapse multimodal, compositional queries into a single embedding and retrieve query-irrelevant frames; HiMu instead parses the query into a logic tree whose leaves (ASR, CLIP, \dots) are scored by modality-specific experts and composed into a per-frame satisfaction curve $T(t)$.
  \textbf{Right:} on Video-MME, HiMu Pareto-dominates prior controlled selectors (AKS, VSLS, BOLT, \revised{T$^\ast$,} MDP3) and uniform sampling across frame budgets, matching uniform-at-$K{=}64$ with only $K{=}16$ frames.}%
  \vspace{0.1cm}%
  }
\g@addto@macro\@maketitle{\himuTeaserFigure}

\newcommand{\himuNotice}{%
  \AddToShipoutPicture*{\AtPageLowerLeft{%
    \put(\LenToUnit{\dimexpr 1in+\oddsidemargin\relax},\LenToUnit{0.4in}){%
      \parbox[b]{\textwidth}{\centering\normalfont\scriptsize\color{black}%
        Accepted to the IEEE/CVF Winter Conference on Applications of Computer Vision (WACV) 2027.\\
        \copyright~2027 IEEE. Personal use of this material is permitted. Permission from IEEE must be obtained for all other uses, in any current or future media, including reprinting/republishing this material for advertising or promotional purposes, creating new collective works, for resale or redistribution to servers or lists, or reuse of any copyrighted component of this work in other works.}}}}}
\makeatother

\graphicspath{{figures/}}

\begin{document}

\addtocontents{toc}{\protect\setcounter{tocdepth}{-1}}
\himuNotice

\title{HiMu: Hierarchical Multimodal Frame Selection\\for Long Video Question Answering}

\author{Dan Ben-Ami$^{1}$ \quad Gabriele Serussi$^{1}$ \quad Kobi Cohen$^{2}$ \quad Chaim Baskin$^{1,3}$\\
$^{1}$INSIGHT Lab, Ben-Gurion University of the Negev, Israel\\
$^{2}$Ben-Gurion University of the Negev, Israel\\
$^{3}$Data Science Research Center, Ben-Gurion University of the Negev, Israel}

\maketitle
\thispagestyle{empty}

\input{sections/00_abstract}

\input{sections/01_intro}
\input{sections/02_related}
\input{sections/03_method}
\input{sections/04_experiments}
\input{sections/05_conclusion}

{\small
\bibliographystyle{ieeenat_fullname}
\bibliography{main}
}

\twocolumn[{\centering\Large\bfseries Appendix\par}\vspace{1.5em}]
\renewcommand{\baselinestretch}{1}\selectfont
\makeatletter
\renewcommand\paragraph{\@startsection{paragraph}{4}{\z@}%
  {3.25ex \@plus1ex \@minus.2ex}%
  {-1em}%
  {\normalfont\normalsize\bfseries}}
\makeatother
\setlength{\textfloatsep}{20pt plus 2pt minus 4pt}
\setlength{\dbltextfloatsep}{20pt plus 2pt minus 4pt}
\renewcommand{\dbltopfraction}{0.7}
\renewcommand{\dblfloatpagefraction}{0.5}
\setcounter{totalnumber}{10}
\setcounter{topnumber}{5}
\setcounter{bottomnumber}{5}
\renewcommand{\topfraction}{.9}
\renewcommand{\bottomfraction}{.9}
\renewcommand{\textfraction}{.05}
\renewcommand{\floatpagefraction}{.95}
\addtocontents{toc}{\protect\setcounter{tocdepth}{1}}

\input{sections/appendix}

\end{document}

%% file: sections/00_abstract.tex
\begin{abstract}
Long-form video question answering requires reasoning over extended temporal contexts, making frame selection a critical bottleneck for multi-modal large language models (MLLMs) bound by finite context windows. Within the controlled frame-budget regime that governs practical deployment, prior selectors score frames against a single global query embedding; as a result, compositional multimodal questions that involve temporal ordering or cross-modal cues such as ``what happens on screen right after the narrator mentions the reaction?'' are flattened into a representation that loses sub-event ordering and modality bindings. We introduce \textbf{HiMu}, a training-free framework for compositional multimodal frame selection. A single text-only LLM call decomposes the query into a hierarchical logic tree whose leaves are atomic predicates, each routed to a lightweight expert spanning vision (CLIP, open-vocabulary detection, OCR) and audio (speech recognition and non-speech sound matching). Expert signals are normalized, smoothed to align across modalities, and composed bottom-up through fuzzy-logic operators that enforce temporal sequencing and adjacency, yielding a continuous per-frame satisfaction curve. Under the standard 16-frame budget on Video-MME, LongVideoBench, and HERBench-Lite, HiMu achieves state-of-the-art accuracy among frame selection methods and improves over uniform sampling across seven diverse MLLMs as a drop-in module, matching the accuracy of uniform sampling at $4\times$ its frame budget, without retraining and without multiple iterative MLLM calls during selection. \revised{Project page: \href{https://danbenami.github.io/HiMu.io/}{\raisebox{-0.15em}{\includegraphics[height=0.95em]{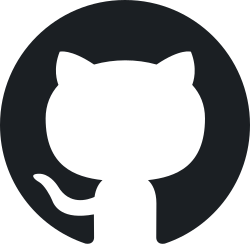}}}}

\end{abstract}

%% file: sections/01_intro.tex
\section{Introduction}
\label{sec:intro}

Long-form video question answering (VideoQA) requires reasoning over extended temporal horizons. Due to the finite context windows of current MLLMs~\cite{qwen3vl2025, internvl, gpt4o}, processing entire videos at native frame rates is computationally infeasible. Practical systems must therefore operate within a controlled frame budget: a small, fixed set of frames passed to the MLLM. Within this regime, frame selection\revised{, a form of visual token reduction~\cite{kong2025token},} becomes a critical bottleneck: regardless of its expressiveness, a model can answer correctly only when given the relevant visual evidence.

The dominant family of controlled frame selectors scores each candidate frame against the query through a frozen vision-language encoder and keeps the top-$K$~\cite{radford2021clip, siglip2024, bolt2025, aks2025, mdp3_2025}. While computationally lightweight, these methods collapse long, compositional queries into a single dense vector, forcing distinct temporal and semantic constraints to be evaluated through one global similarity score. Consider \emph{``After the narrator mentions the chemical reaction, what happens to the beaker on the left?''}: answering it requires reasoning jointly over audio (the narration) and vision (the beaker's state) \emph{and} enforcing their temporal order, but a vision-only encoder is blind to audio and a global similarity score cannot enforce sub-event ordering. \revised{A second line of work injects explicit relational operators into selection~\cite{tstar2025, vsls2025}, using compact, task-specific relation vocabularies or dense per-frame grounding.} Multi-call methods that repeatedly invoke an MLLM within the selection loop~\cite{fan2024videoagent, wang2024videoagent, lvagent2025, longvideoagent2025, sevila2023} reason more richly but \revised{sit in a different compute class (Table~\ref{tab:method_comparison_selector_only}).}

\revised{What is missing is a selector that is at once modality-complete and compositional, yet stays within the controlled frame-budget cost.}
We introduce \textbf{HiMu} (\textbf{Hi}erarchical \textbf{Mu}ltimodal Frame Selection), a training-free framework that resolves compositional and multimodal structure within the controlled frame-budget setting, \emph{without any MLLM call during selection}. The core insight is that complex natural-language queries naturally decompose into structured logic trees. HiMu uses a single text-only LLM call to parse the query into a hierarchical tree of atomic predicates; each leaf is routed to a lightweight modality-specific expert spanning visual experts (CLIP, open-vocabulary detection (OVD), and OCR) and audio experts (automatic speech recognition (ASR) and contrastive language-audio pretraining (CLAP)). The localized signals are evaluated over the timeline and composed bottom-up through continuous fuzzy-logic operators, yielding a per-frame satisfaction curve $T(t) \in [0,1]$ from which the final frames are sampled. \revised{Because the lone LLM call sees only the text query, HiMu avoids the thousands of visual tokens that dominate every MLLM-based selector, keeping per-query cost within the latency class of competing controlled selectors.}

Across Video-MME~\cite{fu2024videomme}, LongVideoBench~\cite{wu2024longvideobench}, and HERBench-Lite~\cite{benami2025herbench}, HiMu achieves state-of-the-art accuracy among controlled frame selection methods. As a plug-and-play module, it consistently improves over uniform sampling for seven diverse MLLMs, ranging from open-source 8B models to proprietary frontier models, without any model-specific tuning. These gains are strongest on longer videos, indicating that query-aware evidence selection can recover relevant context more efficiently than simply increasing uniform coverage.

\noindent Our \textbf{contributions} can be summarized as follows:
\begin{itemize}
    \item A \textbf{\revised{compositional} framework for query-aware frame selection} that decomposes queries into hierarchical logic trees, routes atomic predicates to modality-specific experts, and is the first controlled frame selector to treat audio (ASR and CLAP) as first-class evidence, composing their signals via fuzzy-logic operators enforcing temporal sequencing and adjacency.
    \item A \textbf{training-free, single-shot pipeline} that replaces iterative MLLM inference with one text-only LLM planning step and selectively reused lightweight expert evidence, keeping per-query latency within the class of competing controlled selectors and avoiding MLLM calls during selection.
    \item \textbf{State-of-the-art accuracy among controlled frame selectors} on Video-MME~\cite{fu2024videomme}, LongVideoBench~\cite{wu2024longvideobench}, and HERBench-Lite~\cite{benami2025herbench}, with plug-and-play gains across seven MLLMs and consistent advantages across frame budgets, providing evidence that compositional query representation, not larger context alone, is the missing ingredient.
\end{itemize}

%% file: sections/02_related.tex
\section{Related Work}
\label{sec:related}

\begin{table*}[!t]
\centering
\caption{Comparison of frame selection methods for long-form video QA.
Latency columns report selector-side first-query and cached per-query cost for
a 10-minute video (600 frames at 1\,FPS, $K{=}16$), one query, on
8$\times$ NVIDIA RTX 6000 Pro GPUs, excluding the final QA call.
\revised{HiMu latencies are measured; all others are estimates from reported
numbers normalized to this protocol, and are order-of-magnitude.}
\textcolor{gray}{Gray rows} denote agentic/multi-call methods shown for latency reference only.}
\label{tab:method_comparison_selector_only}
\resizebox{0.9\textwidth}{!}{%
\setlength{\tabcolsep}{8pt}%
\begin{tabular}{l c l c cc c}
\toprule
\makecell[b]{\textbf{Method (Venue)}}
  & \makecell{\textbf{Train-}\\\textbf{Free}}
  & \makecell[b]{\textbf{Query}\\\textbf{Representation}}
  & \makecell[b]{\textbf{Selection}\\\textbf{Evidence}}
  & \makecell[b]{\textbf{First-query}\\\textbf{latency (s)}}
  & \makecell[b]{\textbf{Per-query}\\\textbf{latency (s)}}
  & \makecell[b]{\textbf{Interpretability}\\\textbf{(Compositional Qs)}} \\
\midrule
FFS (CVPR'25)
  & \xmark & Learned policy & Vis.
  & 0.4 & 0.2 & None \\
VidF4 (NLPCC'25)
  & \xmark & Learned score fn. & Vis.
  & 1.0 & 0.3 & None \\
BOLT (CVPR'25)
  & \cmark & Global embedding & Vis.
  & 1.2 & 0.1 & Per-frame score \\
AKS (CVPR'25)
  & \cmark & Global embedding & Vis.
  & 2.0 & 0.2 & Per-frame score \\
MDP$^3$ (ICCV'25)
  & \cmark & Global embedding  & Vis.
  & 2.0 & 0.2 & Per-frame score  \\
\midrule
T$^\ast$ (CVPR'25)
  & \cmark & Flat object queries & Vis.\ (OVD)
  & 7.5 & 7.5 & Detection logs \\
VSLS (NeurIPS'25)
  & \cmark & Fixed relation triplets & Vis.\ (OVD)
  & 8.0 & 7.0 & Detection logs \\
\midrule
\rowcolor[gray]{0.92}
\textbf{HiMu (ours)}
  & \cmark & Hierarchical logic tree & Vis.+Aud.
  & \textbf{4.6} & \textbf{1.9}
  & Frame$\times$Expert scores \\
\midrule
\multicolumn{7}{l}{\color{gray}\textit{\small Agentic / multi-call methods shown for latency reference only; not part of the controlled frame-budget comparison}} \\
\rowcolor{gray!20} \color{gray}VideoZoomer (ICLR'26)~\cite{ding2026videozoomer}
  & \color{gray}\xmark & \color{gray}Implicit (iterative VLM) & \color{gray}Vis.
  & \color{gray}15 & \color{gray}13 & \color{gray}None \\
\rowcolor{gray!20} \color{gray}LVAgent (ICCV'25)~\cite{lvagent2025}
  & \color{gray}\xmark & \color{gray}Implicit (multi-agent MLLM) & \color{gray}Vis.
  & \color{gray}34 & \color{gray}30 & \color{gray}None \\
\rowcolor{gray!20} \color{gray}VCA (ICCV'25)~\cite{yang2025vca}
  & \color{gray}\cmark & \color{gray}Implicit (iterative VLM) & \color{gray}Vis.
  & \color{gray}${\sim}$60 & \color{gray}${\sim}$55 & \color{gray}None \\
\bottomrule
\end{tabular}%
}
\vspace{-0.3cm}
\end{table*}

Controlled frame selection methods for long-video QA aim to identify informative frames within a fixed budget, without invoking the downstream MLLM during selection. \revised{Table~\ref{tab:method_comparison_selector_only} compares these methods with HiMu, which represents the query as an explicit logic tree over both visual and audio evidence.}

\paragraph{Similarity-based frame selection.}
The most efficient selectors score each frame against the query through a frozen vision-language encoder.
BOLT~\cite{bolt2025} pairs query-frame similarity (e.g., CLIP ~\cite{radford2021clip}/ SigLIP~\cite{siglip2024}) with inverse-transform sampling to prioritize relevant frames while preserving selection diversity; AKS~\cite{aks2025} recursively splits the timeline, allocating more keyframes to high-scoring segments; and MDP$^3$~\cite{mdp3_2025} formulates selection as a determinantal point process solved via dynamic programming, capturing relevance, diversity, and sequentiality.
These methods add minimal overhead, yet they collapse multi-clause queries into a single dense representation, offering limited capacity to preserve sub-event ordering or cross-modal bindings that a query may demand (\eg distinguishing a spoken reference from a visual action \emph{and} enforcing their temporal adjacency).
Learning-based variants (e.g. Frame-Voyager~\cite{framevoyager2024}, FFS~\cite{ffs2025}, MLLM-FS~\cite{mllmfs2025}, VidF4~\cite{vidf4_2024}) train scoring or policy modules for richer signals, but require task-specific supervision.
Most selectors in this family compute scores from visual-only features; audio, when available, is typically consumed downstream rather than used to drive selection.

\paragraph{Structured and logic-based selection.}
A second line of work injects explicit relational or logical operators into selection.
T*~\cite{tstar2025} (detector-based variant) casts temporal search as spatial search, using YOLO-World~\cite{yoloworld2024} in an iterative loop that samples frames, detects objects, and reweights the distribution toward relevant evidence.
VSLS~\cite{vsls2025} extends this idea with four predefined dependencies (spatial co-occurrence, temporal proximity, attribute dependency, causal order), using relation checks to iteratively refine the sampling distribution.
\revised{These methods establish that explicit relational structure aids selection, though over flat queries grounded by a single detector family and refined by repeated query-conditioned passes. HiMu extends this to nested composition over heterogeneous visual and audio experts.}
HiMu instead parses the query once into a hierarchical tree and composes cached or query-conditioned expert signals in a non-iterative selection pass, avoiding MLLM calls during selection.

\paragraph{Agentic and multi-call VideoQA systems.}
A separate family of VideoQA systems trades compute for deeper reasoning by repeatedly invoking an MLLM or VLM inside the selection loop, either through agent planners that call tools iteratively~\cite{fan2024videoagent, wang2024videoagent, lvagent2025, longvideoagent2025, videotree2025, ding2026videozoomer, yang2025vca} or MLLM-in-the-loop scorers~\cite{sevila2023, air2025}. These systems address the broader task of finding answer-relevant evidence in long videos, but their iterative visual-language inference places them in a different compute regime from controlled frame selectors; we therefore include representative methods in Table~\ref{tab:method_comparison_selector_only} only as latency references, not as directly comparable selector baselines.

\begin{figure*}[!t]
    \centering
    \includegraphics[width=0.85\textwidth]{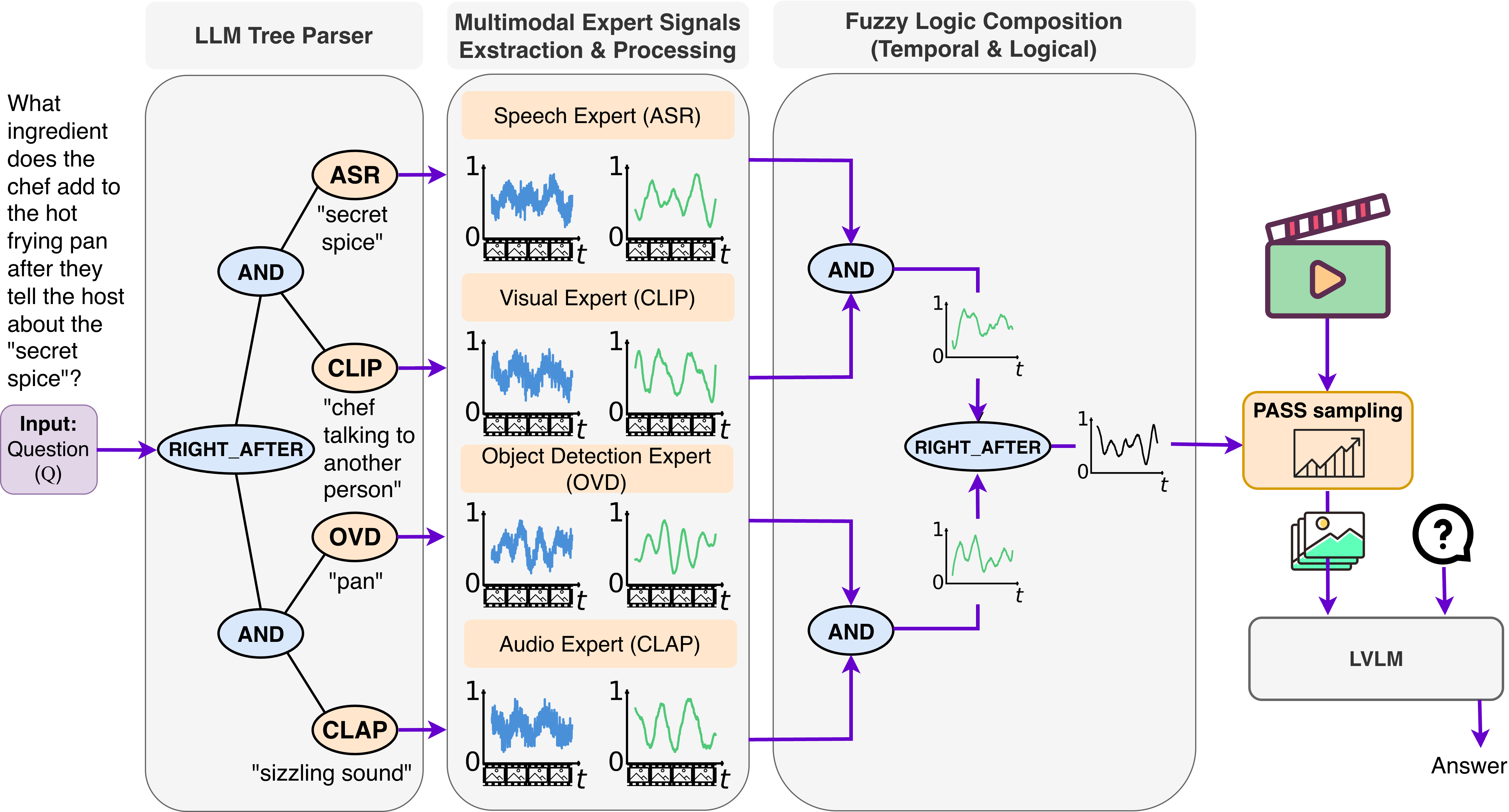}
    \caption{The HiMu pipeline. (1) An LLM parses the question into a logic tree of modality-specific experts. (2) Experts (CLIP, ASR, OVD, CLAP) extract raw signals, which are then normalized and smoothed. (3) Fuzzy logic operators compose signals into a temporal satisfaction curve. (4) Top frames are sampled for the MLLM using PASS.}
    \label{fig:pipeline}
    \vspace{-0.4cm}
\end{figure*}

%% file: sections/03_method.tex
\section{Method}
\label{sec:method}

\revised{We decode the video at a fixed rate (\eg 1\,fps) into $\mathcal{V}=\{v_1,\dots,v_N\}$, where $v_t$ is the RGB frame at timestep $t$ and $a_t$ the audio window centred on it, so every expert signal in Sec.~\ref{ssec:composition} shares the index $t \in \{1,\dots,N\}$.} Given \revised{$\mathcal{V}$ with its aligned audio}, a natural-language question $Q$ (optionally with answer options), and a frame budget \revised{$K \ll N$}, HiMu selects the $K$ most question-relevant frames for a single downstream MLLM call.
The pipeline (Fig.~\ref{fig:pipeline}) proceeds in four stages: (\textit{i})~a text-only LLM decomposes $Q$ into a hierarchical logic tree $\mathcal{T}$ (Sec.~\ref{ssec:parsing}); (\textit{ii})~each leaf is scored by a modality-specific expert, and the resulting signals are lightly post-processed (Sec.~\ref{ssec:experts}); (\textit{iii})~signals are composed bottom-up via fuzzy-logic operators into a per-frame satisfaction curve; and (\textit{iv})~the top-$K$ frames are selected via PASS (Sec.~\ref{ssec:composition}). Thus, the only LLM interaction inside selection is a text-only tree parsing call; expert scoring then proceeds from cached features or from query-conditioned detector (OVD) when the tree requires it.

\subsection{\revised{Hierarchical} Query Decomposition}
\label{ssec:parsing}

A single text-only LLM call receives $Q$ (with answer options, if multiple-choice) and outputs a hierarchical logic tree $\mathcal{T}$ in structured JSON. The tree generation is a text-in, text-out forward pass. The exact system prompt and the JSON schema constraint are provided in \revised{Appendix~\ref{sec:decomposition}}. The tree has two node types:

\paragraph{Leaf nodes.}
Each leaf specifies a modality-specific expert and a text query: $\ell = (\texttt{expert},\,\texttt{query})$, where $\texttt{expert} \in \{\textsc{clip}, \textsc{ovd}, \textsc{ocr}, \textsc{asr}, \textsc{clap}\}$ and \texttt{query} is a natural-language atomic predicate (\eg $\textsc{ovd}$(\texttt{``red car''}) or $\textsc{asr}$(\texttt{``reaction''})). The LLM routes each predicate to the best-suited expert: actions, scenes, and abstract visual concepts $\to$ \textsc{clip}; physical objects and people $\to$ \textsc{ovd}; on-screen text $\to$ \textsc{ocr}; spoken content $\to$ \textsc{asr}; environmental sounds $\to$ \textsc{clap}. Routing rules and worked examples are in the system prompt.

\paragraph{Internal nodes.}
Each internal node applies one of four logical or temporal operators to its children:
\begin{itemize}[leftmargin=*,nosep]
    \item $\textsc{And}$: co-occurrence; all children must be active simultaneously.
    \item $\textsc{Or}$: disjunction; at least one child must be active.
    \item $\textsc{Seq}$: temporal sequence; children are ordered chronologically.
    \item $\textsc{RightAfter}$: tight temporal adjacency; the effect immediately follows the cause.
\end{itemize}

\begin{figure*}[t]
    \centering
    \includegraphics[width=0.85\textwidth]{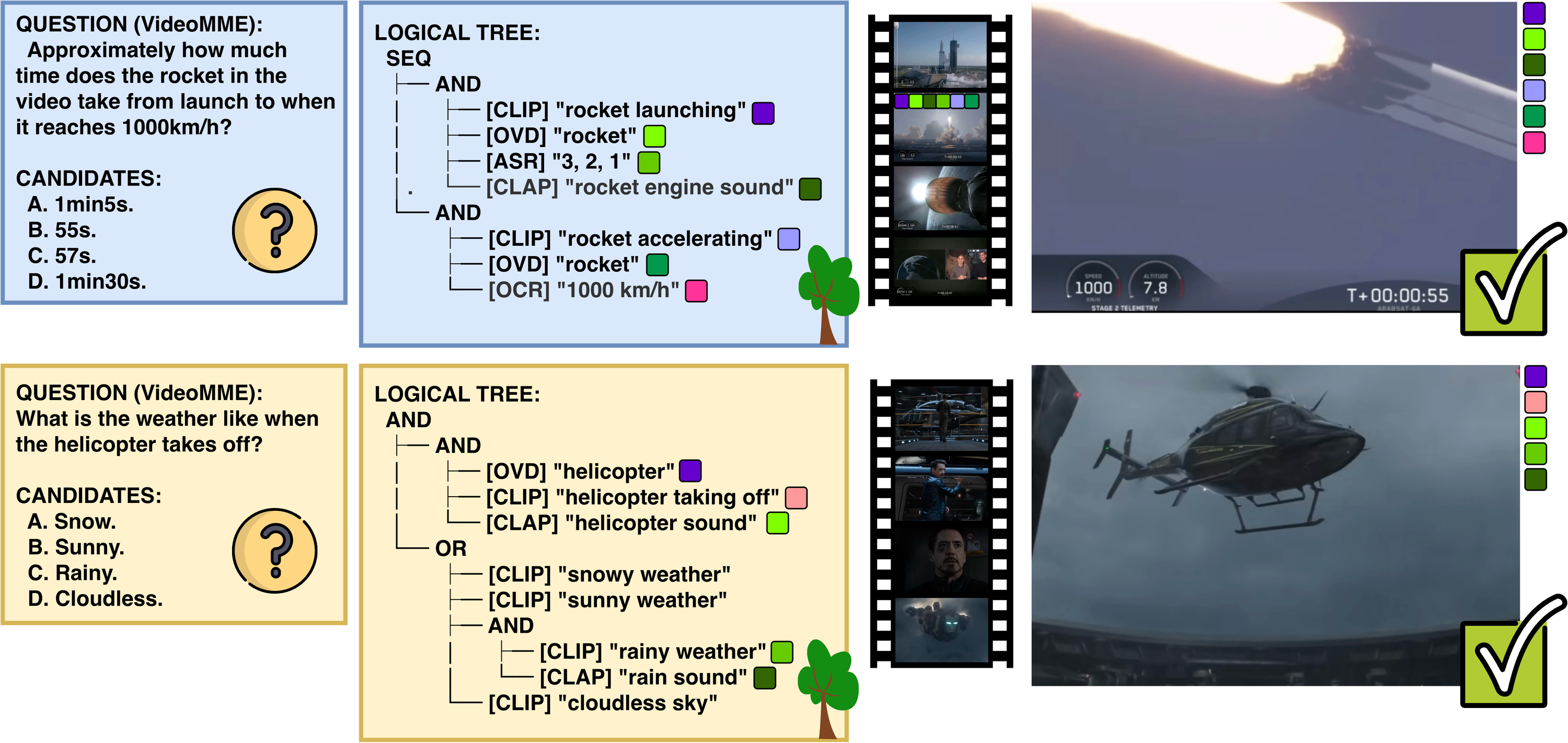}
    \caption{\revised{Logic-tree illustrations for two Video-MME questions, simplified for visual clarity to convey how visual and audio predicates are combined through logical and temporal operators. Coloured chips mark the leaves active on the displayed selected frame, exposing the per-leaf attribution behind each selection; \revised{Appendix~\ref{sec:interpretability}} shows the full per-frame trace.}}
    \label{fig:trees}
    \vspace{-0.4cm}
\end{figure*}

\paragraph{MCQ tree pattern.}
For MCQs, the tree factors shared context from answer-specific branches, typically following the \texttt{And(}\allowbreak\texttt{shared\_context, Or(}\allowbreak\texttt{option\_1, ..., option\_n))} pattern. Each option branch decomposes into expert-specific predicates (Fig.~\ref{fig:trees}).

\subsection{Multimodal Expert Signals Extraction and Processing}
\label{ssec:experts}

Leaf nodes are grouped by expert type for efficient batched inference. Each expert produces a per-frame raw relevance signal $u_i(t) \in \mathbb{R}$ for leaf $i$ at every timestamp $t \in \{1,\dots,N\}$. Five experts span two modality categories; to our knowledge, no prior frame selector leverages audio experts, and we show their inclusion is critical for key-moment discovery.

\noindent\textbf{Visual experts.}
\textbf{CLIP}~\cite{radford2021clip} computes cosine similarity between frame and text-query embeddings, mapped to $[0,1]$; frame embeddings are extracted once and shared across all \textsc{clip} leaves.
\textbf{OVD}~\cite{yoloworld2024} runs open-vocabulary object detection, returning the maximum detection confidence for the queried class per frame; query variations (singular/plural, with/without adjectives) are generated for robust matching.
\textbf{OCR}~\cite{doctr2021} performs on-screen text recognition with substring and Levenshtein-distance fuzzy matching.

\noindent\textbf{Audio experts.}
\textbf{ASR}~\cite{whisper2023} transcribes the audio track once into timestamped word segments; queries are matched via exact substring matching (score $=1.0$) or, failing that, semantic similarity via a sentence-embedding model, with segment scores mapped to frames by temporal-overlap weighting.
\textbf{CLAP}~\cite{clap2023} computes cosine similarity between frame-aligned audio chunks and the text query for non-speech sounds (environmental sounds, effects, music).

\noindent\textbf{Caching and conditional execution.}
CLIP, ASR, CLAP, and OCR features are query-independent and cached per video\revised{, following index-scale video retrieval~\cite{wagner2026vedje}}; only OVD is query-conditioned and re-run per query. Unused experts are skipped entirely.

\paragraph{Normalization.}
Raw expert scores live on incomparable scales (CLIP cosine similarities, OVD confidences, binary ASR matches), each with different ranges and noise profiles. Each signal $u_i$ is mapped to $(0,1)$ via:
\vspace{-0.1cm}
\begin{equation}
    \tilde{u}_i(t) \;=\; \sigma\!\left(\gamma \cdot \frac{u_i(t) - \mathrm{med}(u_i)}{\mathrm{MAD}(u_i) + \delta}\right),
    \label{eq:norm}
\end{equation}
where $\mathrm{med}(\cdot)$ and $\mathrm{MAD}(\cdot)$ denote the median and Median Absolute Deviation, $\sigma(\cdot)$ is the sigmoid, $\gamma$ controls sharpness, and $\delta$ is a small stabilizer. Median/MAD gives robustness to the heavy-tailed distributions typical of detection and retrieval models, and the sigmoid yields a smooth, fuzzy-logic-compatible mapping. When multiple leaves share an expert, statistics are computed jointly over all their signals, so a confident detection and a weak one are not stretched alike to $[0,1]$ and falsely implied to be equally relevant.

\paragraph{Bandwidth-matched smoothing.}
After normalization, each signal is convolved with a modality-specific Gaussian kernel:
\vspace{-0.1cm}
\begin{equation}
    \hat{u}_i(t) \;=\; \sum_{t'=1}^{N} \tilde{u}_i(t')\;\mathcal{G}\!\left(t - t';\;\sigma_{m}\right),
    \label{eq:smooth}
\end{equation}
where $\mathcal{G}(\Delta;\sigma) = \frac{1}{\sqrt{2\pi}\,\sigma}\exp\!\big({-\Delta^2}/{2\sigma^2}\big)$ and $m = \texttt{expert}_i$. Visual signals (CLIP, OVD, OCR) are frame-precise and receive narrow kernels; ASR and CLAP have coarser temporal resolution and receive wider kernels. This resolves cross-modal asynchrony by ensuring that peaks from different modalities overlap temporally, preventing missed conjunctions at the composition stage.

\subsection{Fuzzy Logic Composition}
\label{ssec:composition}

\paragraph{Bottom-up tree evaluation.}
The logic tree $\mathcal{T}$ is evaluated bottom-up: leaf nodes return their processed signals $\hat{u}_i(t)$; internal nodes apply continuous fuzzy-logic operators. We use four operators as a compact vocabulary for VideoQA evidence: co-occurrence, alternatives, ordered events, and tight ``right after'' transitions.

\noindent\textbf{Logical operators.} Logical composition follows standard fuzzy operators that keep scores in $[0,1]$: the product t-norm suppresses \textsc{And} when any child is weak, while the probabilistic sum lets either branch satisfy \textsc{Or} without unbounded accumulation. \revised{Signals enter each operator rescaled to $[0.5,1]$, so repeated conjunction attenuates gracefully over the shallow trees the parser produces (mean depth $2.71{\pm}0.44$).} Both are applied pairwise left-to-right for $n{>}2$ children:
\vspace{-0.1cm}
\begin{align}
    \textsc{And}(A,B)(t) &= A(t) \cdot B(t), \label{eq:and} \\[3pt]
    \textsc{Or}(A,B)(t) &= A(t) + B(t) - A(t) \cdot B(t). \label{eq:or}
\end{align}

\noindent\textbf{Temporal operators.}

\noindent$\textsc{Seq}$ \emph{(temporal ordering).}
Given children $(u_1,\dots,u_L)$ in chronological order, this operator enforces sequence through past/future evidence rather than brittle timestamp alignment, while still selecting frames from \emph{every} step:
\begin{equation}
    \textsc{Seq}(t) \;=\; \max_{\ell \in \{1,\dots,L\}} \left[\, u_\ell(t) \;\cdot \prod_{j<\ell} H_j(t) \;\cdot \prod_{j>\ell} F_j(t) \,\right],
    \label{eq:seq}
\end{equation}
where $H_j(t) = \max_{s < t}\, u_j(s)$ is the \emph{has-occurred} signal (running max up to $t$) and $F_j(t) = \max_{s > t}\, u_j(s)$ is the \emph{yet-to-occur} signal (running max after $t$). Thus, a step $\ell$ activates at time $t$ only when earlier steps have already peaked and later steps still peak in the future; the outer max lets all events contribute, not just the final one.

\noindent$\textsc{RightAfter}$ \emph{(tight temporal proximity).}
For event pairs that should happen close together in time, this operator uses exponential decay rather than a fixed window: a frame scores highly when the paired event occurred nearby, with sharpness controlled by $\kappa$:
\vspace{-0.1cm}
\begin{align}
    S_\text{effect}(t) &= \text{effect}(t) \cdot {\textstyle\sum_{s<t}} \text{cause}(s)\, e^{-\kappa(t-s)}, \label{eq:ra_eff} \\[3pt]
    S_\text{cause}(t) &= \text{cause}(t) \cdot {\textstyle\sum_{s>t}} \text{effect}(s)\, e^{-\kappa(s-t)}, \label{eq:ra_cau} \\[3pt]
    \textsc{RightAfter}(t) &= \max\!\big(S_\text{effect}(t),\; S_\text{cause}(t)\big). \label{eq:ra}
\end{align}
The two terms ensure frames are selected from both the cause and the effect side: $S_\text{effect}$ scores effect frames weighted by how recently the cause fired, while $S_\text{cause}$ does the reverse.

The root of $\mathcal{T}$ produces the \textbf{satisfaction curve} $T(t) \in [0,1]$, a per-frame composite score reflecting how well the entire logic tree is satisfied at each timestamp.

\paragraph{PASS: Peak-And-Spread Selection.}
\label{ssec:selection}
Na\"ively selecting the top-$K$ frames of $T(t)$ over-concentrates on a single high-scoring segment, missing other relevant events and short-term motion context. We therefore introduce \textbf{PASS} (\textbf{P}eak-\textbf{A}nd-\textbf{S}pread \textbf{S}election): we first select $N_p$ local maxima of $T(t)$ with a minimum inter-peak distance $\Delta$, preventing redundant peaks without forcing coverage of low-satisfaction regions. Each peak is augmented with its $N_n$ highest-scoring neighbors within a window of size $w$, capturing short-term motion while scaling local context with the budget. The remaining budget is filled greedily from the highest-scoring unselected frames of $T(t)$.
\revised{All four quantities are fixed from the budget, $N_p = w = \Delta = \lfloor\sqrt{K}\rfloor$ and $N_n = \lfloor\sqrt{K}/2\rfloor$, with no benchmark-specific tuning.}
\revised{At $K{=}16$, PASS improves over vanilla top-$K$ by $0.86$\,pp on Video-MME. Independently halving or doubling each parameter changes accuracy by at most $0.72$\,pp, indicating that the improvement is robust around the derived setting.}
Each selected frame $v_{t^*}$ carries its per-leaf scores $\{\hat{u}_i(t^*)\}$, an interpretable trace of \emph{which experts and predicates} drove its selection.
\revised{Detailed pseudocode, the parameter sweep, and per-leaf heatmap examples illustrating HiMu's interpretability are provided in Appendices~\ref{sec:pass}, \ref{ssec:pass_sensitivity}, and~\ref{sec:interpretability}.}

%% file: sections/04_experiments.tex
\section{Experiments}
\label{sec:experiments}

We evaluate HiMu along \revised{six} axes: controlled accuracy against prior frame selectors, plug-and-play gains across downstream MLLMs, the contribution of the compositional design, behavior under different frame budgets, \revised{the reliability of the generated logic trees,} and selector latency.

\subsection{Setup}
\label{subsec:setup}

\paragraph{Benchmarks.}
We use three complementary long-video QA benchmarks.
\textbf{Video-MME}~\cite{fu2024videomme} contains 2,700 multiple-choice questions over 900 videos, split into Short ($<$2\,min), Medium (4--15\,min), and Long (30--60\,min) durations; its native audio and duration splits test both multimodal grounding and temporal scale.
\textbf{LongVideoBench$_\text{val}$}~\cite{wu2024longvideobench} contains roughly 1.3K validation questions with referring queries over 17 categories, where subtitles or Whisper transcripts serve as speech evidence.
\textbf{HERBench-Lite}~\cite{benami2025herbench} contains 2K purely visual questions requiring integration of at least three non-overlapping cues, stressing whether selected frames can support multi-evidence reasoning once audio is unavailable.

\vspace{-0.2cm}
\paragraph{Implementation details.}
Unless stated otherwise, all methods select $K{=}16$ frames from videos sampled at 1\,fps and answer with the same downstream MLLM within each comparison.
HiMu is training-free: the logic tree is generated by the same LLM used as the answerer.
The default experts are CLIP-dfn~\cite{dfn2024}, YOLO-World v2~\cite{yoloworld2024} (OVD), docTR~\cite{doctr2021} (OCR), faster-whisper large-v3-turbo~\cite{whisper2023} (ASR), and LAION CLAP~\cite{clap2023}; all reusable features except query-conditioned OVD are cached per video.
All hyperparameters are fixed once across benchmarks, models, and ablations with no per-dataset tuning. \revised{Appendices~\ref{sec:implementation}, \ref{sec:sensitivity}, and~\ref{sec:latency}} report the exact values, sensitivity analyses, backbone and parser swaps, and the latency breakdown.

\paragraph{Baselines.}
The controlled comparison fixes the answerer to Qwen3-VL-8B~\cite{qwen3vl2025} and compares HiMu to uniform sampling, similarity-based selectors BOLT~\cite{bolt2025}, AKS~\cite{aks2025}, and MDP$^3$~\cite{mdp3_2025}, and structured selectors T$^\ast$~\cite{tstar2025} and VSLS~\cite{vsls2025}.
\revised{All reported baseline numbers are our reproductions using the authors' released implementations under the common Qwen3-VL-8B answerer, 1\,fps candidate pool, and $K{=}16$ frame budget. We additionally report Random Sampling, which draws $K$ frames without replacement from the same candidates and is averaged over three seeds (mean $\pm$ std).}
The selector inputs follow each benchmark's modality regime: on Video-MME, frame selectors operate without subtitles, using the video/audio stream; on LongVideoBench$_\text{val}$, audio is unavailable, so selectors may use the benchmark-provided subtitles as the speech-language stream; HERBench-Lite is visual-only.
At answer time, the downstream MLLM always receives the same selected frames, question, candidates, and full subtitle/transcript context for Video-MME and LongVideoBench$_\text{val}$, since many questions require speech evidence.
This keeps the selector as the only variable within each benchmark: performance differences reflect which visual evidence the selector surfaces rather than whether text evidence is withheld.

\subsection{Main Results}
\label{subsec:main_results}

\addtocounter{table}{1}
\begin{table*}[!t]
    \centering
    \caption{Generalization of HiMu as a plug-and-play frame selector across seven diverse MLLMs at $K{=}16$ frames. HiMu delivers broad gains over uniform sampling, with the strongest improvements on Video-MME and LongVideoBench$_\text{val}$.}
    \label{tab:main_results_generalization}
    \resizebox{0.86\textwidth}{!}{
    \setlength{\tabcolsep}{8pt}
    \begin{tabular}{l l cccc cc}
        \toprule
        \multirow{2.5}{*}{\textbf{Method}} & \multirow{2.5}{*}{\textbf{Model}} &
        \multicolumn{4}{c}{\textbf{Video-MME}} & \multirow{2.5}{*}{\textbf{LVB$_\text{val}$}} &
        \multirow{2.5}{*}{\textbf{HERBench-Lite}} \\
        \cmidrule(lr){3-6}
        & & \textbf{Short} & \textbf{Medium} & \textbf{Long} & \textbf{Overall} & & \\
        \midrule
        Uniform & Qwen3-VL-8B & 76.3 & 66.3 & 55.6 & 66.1 & 55.7 & 41.7 \\
        \rowcolor{gray!10} \textbf{HiMu (Ours)} & Qwen3-VL-8B & \textbf{78.6} & \textbf{71.0} &
        \textbf{69.9} & \textbf{73.2} & \textbf{64.2} & \textbf{43.2} \\
        \midrule
        Uniform & LLaVA-OV-1.5-8B & \textbf{72.3} & 62.3 & 54.9 & 63.2 & 54.3 & 35.8 \\
        \rowcolor{gray!10} \textbf{HiMu (Ours)} & LLaVA-OV-1.5-8B & 71.9 & \textbf{67.0} &
        \textbf{63.9} & \textbf{67.6} & \textbf{57.9} & \textbf{35.9} \\
        \midrule
        Uniform & InternVL-3.5-8B & 75.4 & 67.4 & 56.6 & 66.6 & 59.2 & 38.3 \\
        \rowcolor{gray!10} \textbf{HiMu (Ours)} & InternVL-3.5-8B & \textbf{76.9} & \textbf{70.4} &
        \textbf{66.5} & \textbf{71.4} & \textbf{64.1} & \textbf{38.3} \\
        \midrule
        Uniform & Qwen2.5-VL-7B & 72.5 & 61.0 & 53.8 & 62.6 & 54.6 & 34.1 \\
        \rowcolor{gray!10} \textbf{HiMu (Ours)} & Qwen2.5-VL-7B & \textbf{73.1} & \textbf{65.1} &
        \textbf{62.9} & \textbf{67.1} & \textbf{57.5} & \textbf{35.2} \\
        \midrule
        Uniform & Gemma-3-12B & \textbf{73.3} & 60.3 & 55.8 & 63.0 & 47.6 & 31.2 \\
        \rowcolor{gray!10} \textbf{HiMu (Ours)} & Gemma-3-12B & 71.6 & \textbf{65.7} &
        \textbf{67.5} & \textbf{68.3} & \textbf{53.9} & \textbf{31.5} \\
        \midrule
        \multicolumn{8}{l}{\textit{Proprietary MLLMs (evaluated on a stratified random 25\% subset of each benchmark)}} \\
        Uniform & Gemini-2.5-Flash & 78.4 & 67.1 & 61.2 & 69.0 & 56.3 & 37.3 \\
        \rowcolor{gray!10} \textbf{HiMu (Ours)} & Gemini-2.5-Flash & \textbf{78.9} & \textbf{75.0} &
        \textbf{74.5} & \textbf{76.1} & \textbf{70.1} & \textbf{37.7} \\
        \midrule
        Uniform & GPT-4o & 77.0 & 73.0 & 71.4 & 73.8 & 55.6 & 37.5 \\
        \rowcolor{gray!10} \textbf{HiMu (Ours)} & GPT-4o & \textbf{80.9} & \textbf{77.2} &
        \textbf{76.5} & \textbf{78.2} & \textbf{65.1} & \textbf{40.7} \\
        \bottomrule
    \end{tabular}
    }
    \vspace{-0.3cm}
\end{table*}
\addtocounter{table}{-2}

\begin{table}[t]
    \centering
    \caption{Controlled comparison of frame selection methods at $K{=}16$ frames with Qwen3-VL-8B. \revised{HiMu-Visual uses the same visual modalities as the baselines (CLIP, OVD, OCR); full HiMu additionally uses each benchmark's speech/audio evidence.}}
    \label{tab:main_results_controlled}
    \resizebox{\linewidth}{!}{
    \setlength{\tabcolsep}{4.5pt}
    \begin{tabular}{l cccc cc}
        \toprule
        \multirow{2.5}{*}{\textbf{Method (Venue)}} &
        \multicolumn{4}{c}{\textbf{Video-MME}} & \multirow{2.5}{*}{\textbf{LVB$_\text{val}$}} &
        \multirow{2.5}{*}{\textbf{HERBench-Lite}} \\
        \cmidrule(lr){2-5}
        & \textbf{Short} & \textbf{Medium} & \textbf{Long} & \textbf{Overall} & & \\
        \midrule
        \revised{Random Sampling} & \revised{$74.7{\pm}0.2$} & \revised{$64.7{\pm}0.3$} & \revised{$53.6{\pm}0.3$} & \revised{$64.4{\pm}0.3$} & \revised{$55.8{\pm}0.3$} & \revised{$41.4{\pm}0.1$} \\
        Uniform Sampling & 76.3 & 66.3 & 55.6 & 66.1 & 55.7 & 41.7 \\
        BOLT~\cite{bolt2025} (CVPR'25) & 69.6 & 67.9 & 68.7 & 68.7 & 54.6 & 42.2 \\
        T$^\ast$~\cite{tstar2025} (CVPR'25) & 73.7 & 67.4 & 68.1 & 69.8 & 57.5 & 39.1 \\
        AKS~\cite{aks2025} (CVPR'25) & 70.1 & 65.1 & 68.7 & 68.0 & 57.1 & 40.3 \\
        MDP$^3$~\cite{mdp3_2025} (ICCV'25) & 75.4 & 62.4 & 64.7 & 67.5 & 55.6 & 37.5 \\
        VSLS~\cite{vsls2025} (NeurIPS'25) & 74.1 & 66.7 & 69.2 & 70.0 & 58.7 & 39.7 \\
        \midrule
        HiMu-Visual (ours) & 77.8 & 70.9 & 69.4 & 72.8 & 61.0 & 43.2 \\
        \rowcolor{gray!10} \textbf{HiMu (ours)} & \textbf{78.6} & \textbf{71.0} &
        \textbf{69.9} & \textbf{73.2} & \textbf{64.2} & \textbf{43.2} \\
        \bottomrule
    \end{tabular}
    }
    \vspace{-0.3cm}
\end{table}
\addtocounter{table}{1}

\revon
\noindent\textbf{(Q1) HiMu is the strongest controlled selector in both modality-matched and multimodal comparisons.}
Table~\ref{tab:main_results_controlled} fixes the answerer to Qwen3-VL-8B. In the modality-matched visual-only comparison, \textbf{HiMu-Visual} reparses each query using CLIP, OVD, and OCR and exceeds the strongest visual baseline by $2.8$\,pp on Video-MME, $2.3$\,pp on LongVideoBench$_\text{val}$, and $1.0$\,pp on HERBench-Lite. This isolates the contribution of hierarchical composition from access to additional audio evidence.
Full HiMu reaches $73.2\%$, $64.2\%$, and $43.2\%$, respectively. Relative to HiMu-Visual, speech and non-speech audio evidence adds $0.4$\,pp on Video-MME and $3.2$\,pp on LongVideoBench$_\text{val}$; the two configurations coincide on the visual-only HERBench-Lite. The duration splits further show that the benefit is concentrated where temporal coverage is sparse, with the largest Video-MME margin on Medium and Long videos.
\revoff

\vspace{0.2cm}
\noindent\textbf{(Q2) HiMu transfers as a plug-and-play selector.}
Table~\ref{tab:main_results_generalization} evaluates seven MLLMs under the same HiMu configuration: five open-source models and two proprietary models.
Replacing uniform sampling with HiMu improves Video-MME Overall for every model, with gains from $+4.4$ to $+7.1$pp, and improves LongVideoBench$_\text{val}$ by $+2.9$ to $+13.8$pp.
HERBench-Lite also improves or matches uniform for every model, though margins are naturally smaller because the benchmark removes audio/subtitles and stresses the answerer's multi-frame fusion ability.
Across models, the gains concentrate on Medium and Long Video-MME splits; uniform sampling only occasionally matches or exceeds HiMu on Short videos, where dense temporal coverage is easier. This pattern supports the central claim: HiMu's benefit comes from query-aware evidence selection, not from a special interaction with one MLLM.

\vspace{0.2cm}
\subsection{Component Analysis and Frame Budget}
\label{subsec:ablation}

\noindent\textbf{(Q3) The gains come from hierarchical composition, not simply more signals.}
Table~\ref{tab:expert_ablation} ablates HiMu on Video-MME using Qwen3-VL-8B.
The largest drop comes from replacing the logic tree with Flat Fusion ($-5.5$pp), which pools all leaf scores without hierarchy or typed operators.
This loss is larger than removing any individual expert, showing that the main bottleneck is how evidence is composed.
The structural rows further separate the two ingredients: flattening the tree while preserving the LLM-chosen root operator still costs $-1.4$pp, while operator substitutions cost $-1.0$ to $-2.5$pp.
Conjunction is the most sensitive operator ($\textsc{And}{\to}\textsc{Or}$, $-2.5$pp), but temporal operators also matter ($\textsc{Seq}{\to}\textsc{And}$, $-1.7$pp; $\textsc{RightAfter}{\to}\textsc{And}$, $-1.3$pp). Thus the improvement is not reducible to temporal logic alone; it comes from the joint hierarchy-plus-operator design.

\revon
Expert leave-one-out holds the parsed tree fixed and neutralizes one expert's leaves before composition, isolating that expert's contribution within the original multimodal plan. Under this intervention, ASR has the largest effect ($-2.0$\,pp), followed by CLIP ($-1.4$\,pp), OCR and CLAP ($-1.0$\,pp each), and OVD ($-0.7$\,pp). The answerer receives identical subtitle context in every condition, so these differences arise from frame selection.
HiMu-Visual measures a different, adaptive setting: it removes both audio experts and reparses the query for the remaining visual experts. Its $0.4$\,pp Video-MME gap therefore quantifies the deployable visual-only system rather than the marginal effect of deleting ASR from a fixed multimodal tree. A matched control confirms this distinction: jointly neutralizing ASR and CLAP in the fixed multimodal tree scores $71.1\%$ ($-2.1$\,pp), whereas reparsing with the same visual-only expert set reaches $72.8\%$, recovering $1.7$\,pp through adaptation. Backbone and parser swaps \revised{(Appendix~\ref{sec:sensitivity})} vary by at most $0.6$ and $0.8$\,pp, respectively.
\revoff

\medskip
\noindent\textbf{(Q4) HiMu keeps its advantage across frame budgets.}
Table~\ref{tab:frame_budget} varies $K \in \{8,16,32,64\}$.
HiMu improves Video-MME Overall over uniform sampling at every budget, by $+6.4$, $+7.1$, $+4.9$, and $+4.1$pp respectively.
\revised{The margin peaks at $K{=}16$ and narrows as the budget grows: uniform coverage saturates, so the evidence HiMu recovers is increasingly sampled anyway.}
The gains are concentrated on Medium and Long videos, where uniform sampling is most likely to miss query-relevant moments\revised{, while on the Short split uniform is $0.6$ and $0.5$\,pp higher at $K{=}32$ and $64$}.
\revised{Appendix~\ref{sec:scope} confirms that this boundary is a property of the regime rather than of Video-MME: HiMu gains $0.8$\,pp on NExT-QA, whose videos are short relative to the budget, and trails uniform by $0.9$\,pp on EgoSchema, whose questions are about the video as a whole.}
Most importantly, HiMu with only $K{=}16$ frames (73.2\%) exceeds uniform sampling with $K{=}64$ frames (71.7\%)\revised{, a $4\times$ budget reduction showing} that the bottleneck is not only context length: many uniformly sampled frames are simply not the evidence requested by the question.

\medskip
\vspace{-0.05cm}
\revon
\noindent\textbf{(Q5) Direct audits characterize tree quality.}
Every generated tree is checked at run time against an acceptance test covering schema conformance, the closed operator vocabulary, and the benchmark's modality regime; across the three benchmarks $99.2\%$ of questions are accepted within a bounded retry budget, and the remaining $0.8\%$ receive a flat fallback tree.
Well-formedness is not fidelity, so a human study of $270$ question--tree pairs ($90$ per benchmark; mean pairwise Cohen's $\kappa{=}0.79$) finds $92.2\%$ of trees fully correct.
The same study finds the four operators sufficient for $96.7$--$100\%$ of sampled queries; the remaining cases request absolute temporal position, which composition does not enforce \revised{(Appendix~\ref{ssec:human_eval_analysis})}.
The residual errors are mild in effect: manually correcting the $21$ flagged trees leaves $19$ of $21$ predictions unchanged, with no correct-to-incorrect transition, whereas deliberate corruptions of the tree cost $1.4$--$5.5$\,pp (Table~\ref{tab:expert_ablation}).
Composition quality is therefore what drives the gains, and the parser supplies it reliably.
\revised{Appendices~\ref{sec:parser_reliability} and~\ref{sec:tree_human_eval}} report the full audit protocol, the human-study design, and the per-benchmark breakdowns.
\revoff

\begin{table}[t]
    \centering
    \caption{Component ablation on Video-MME (Qwen3-VL-8B, $K{=}16$). \emph{Compositional structure}: ``Flat Fusion'' pools all leaf scores with no tree or typed operators; ``w/o nesting'' flattens the tree but keeps the root operator; each ``$X{\to}Y$'' row swaps one operator type. \emph{Experts} removes one expert at a time.}
    \label{tab:expert_ablation}
    \setlength{\tabcolsep}{6pt}
    \resizebox{\columnwidth}{!}{
        \begin{tabular}{l cccc}
            \toprule
            \multirow{2.5}{*}{\textbf{Configuration}} & \multicolumn{4}{c}{\textbf{Video-MME}} \\
            \cmidrule(lr){2-5}
            & \textbf{Short} & \textbf{Med.} & \textbf{Long} & \textbf{Overall} \\
            \midrule
            \textbf{HiMu} & \textbf{78.6} & \textbf{71.0} & \textbf{69.9} & \textbf{73.2} \\
            \midrule
            \multicolumn{5}{l}{\textit{Compositional structure}} \\
            Flat Fusion & 68.7 & 66.1 & 68.5 & 67.7 \\
            w/o nesting & 76.9 & 70.6 & 67.6 & 71.8 \\
            \textsc{Or}$\to$\textsc{And} & 77.5 & 70.4 & 68.5 & 72.2 \\
            \textsc{Seq}$\to$\textsc{And} & 76.8 & 69.2 & 68.5 & 71.5 \\
            \textsc{RightAfter}$\to$\textsc{And} & 77.4 & 69.3 & 68.8 & 71.9 \\
            \textsc{And}$\to$\textsc{Or} & 74.0 & 68.4 & 69.7 & 70.7 \\
            \midrule
            \multicolumn{5}{l}{\textit{Experts (leave-one-out)}} \\
            w/o ASR & 76.3 & 69.1 & 68.1 & 71.2 \\
            w/o CLAP & 78.0 & 69.4 & 69.1 & 72.2 \\
            w/o CLIP & 76.5 & 69.6 & 69.2 & 71.8 \\
            w/o OCR & 77.4 & 69.9 & 69.1 & 72.2 \\
            w/o OVD & 77.4 & 70.6 & 69.2 & 72.5 \\
            \bottomrule
        \end{tabular}
    }
    \vspace{-0.35cm}
\end{table}

\begin{table}[t]
    \centering
    \caption{Effect of frame budget~$K$ on Video-MME (Qwen3-VL-8B). HiMu consistently improves Overall and is especially strong on Medium and Long videos.}
    \label{tab:frame_budget}
    \setlength{\tabcolsep}{6pt}
    \resizebox{\columnwidth}{!}{
        \begin{tabular}{c l cccc}
            \toprule
            \multirow{2.5}{*}{\textbf{\# Frames}} & \multirow{2.5}{*}{\textbf{Method}} & \multicolumn{4}{c}{\textbf{Video-MME}} \\
            \cmidrule(lr){3-6}
            & & \textbf{Short} & \textbf{Med.} & \textbf{Long} & \textbf{Overall} \\
            \midrule
            \multirow{2}{*}{8}
            & Uniform & \textbf{73.5} & 63.4 & 54.7 & 64.0 \\
            & HiMu & \textbf{73.5} & \textbf{68.7} & \textbf{68.8} & \textbf{70.4} \\
            \midrule
            \multirow{2}{*}{16}
            & Uniform & 76.3 & 66.3 & 55.6 & 66.1 \\
            & \textbf{HiMu} & \textbf{78.6} & \textbf{71.0} & \textbf{69.9} & \textbf{73.2} \\
            \midrule
            \multirow{2}{*}{32}
            & Uniform & \textbf{81.4} & 69.2 & 58.7 & 69.9 \\
            & HiMu & 80.8 & \textbf{73.3} & \textbf{70.0} & \textbf{74.8} \\
            \midrule
            \multirow{2}{*}{64}
            & Uniform & \textbf{82.9} & 71.1 & 60.6 & 71.7 \\
            & HiMu & 82.4 & \textbf{73.5} & \textbf{71.1} & \textbf{75.8} \\
            \bottomrule
        \end{tabular}
    }
    \vspace{-0.3cm}
\end{table}

\subsection{Efficiency Analysis}
\label{subsec:efficiency}

\noindent\textbf{(Q\revised{6}) HiMu trades modest selector cost for the best controlled accuracy.}
Table~\ref{tab:method_comparison_selector_only} reports standardized selector-only latency for a 10-minute video at 1\,FPS on 8$\times$ NVIDIA RTX 6000 Pro GPUs, excluding the final QA call that every method pays.
Lightweight learned and similarity-based selectors remain fastest (0.4--2.0\,s first-query latency; 0.1--0.3\,s per-query latency) but remain below HiMu in Table~\ref{tab:main_results_controlled}.
HiMu's measured selector latency is 4.6\,s for the first query and 1.9\,s for each cached per-query run.
Structured selectors with heavier query-conditioned grounding, T$^\ast$ and VSLS, require 7.5--8.0\,s first-query latency and 7.0--7.5\,s per-query latency, while VideoZoomer, LVAgent, and VCA are included only as latency references and are slower by a wider margin under the standardized estimates.
\revised{Appendix~\ref{sec:latency}} breaks HiMu's cost down: reusable CLIP/OCR/ASR/CLAP features are cached, while the per-query cost is dominated by tree parsing and OVD.
This is the intended trade-off: HiMu spends a few seconds beyond global visual retrieval to obtain compositional, visual-plus-audio evidence selection, per-frame expert traces, and the best accuracy across all controlled benchmarks.

%% file: sections/05_conclusion.tex
\vspace{-0.1cm}
\section{Discussion}
\label{sec:discussion}

\revon
HiMu shows that the reasoning a compositional query demands need not be paid for in MLLM calls: one text-only planning step and a set of lightweight experts recover the modality bindings and temporal order that a single global similarity score discards. It follows that much of what larger uniform budgets buy is redundant coverage, and that matching a $4\times$ budget at $K{=}16$ is a matter of representing the query well rather than of seeing more frames.

\vspace{-0.4cm}
\paragraph{Limitations.} The design assumes questions that name localizable evidence in videos that are long relative to the frame budget; where coverage is already dense or the answer depends on the video as a whole, uniform sampling is the better cost--accuracy choice. Multiple choice supplies the candidate branches that let selection cover alternatives, including constraints the composition stage does not enforce directly, whereas open-ended questions use the same decomposition without them. HiMu also carries a heavier footprint than a single-encoder selector: five expert models to deploy and a per-query cost above that of similarity-based retrieval.
\revoff

%% file: sections/appendix.tex
\begin{figure*}[t]
  \centering
  \includegraphics[width=0.92\textwidth]{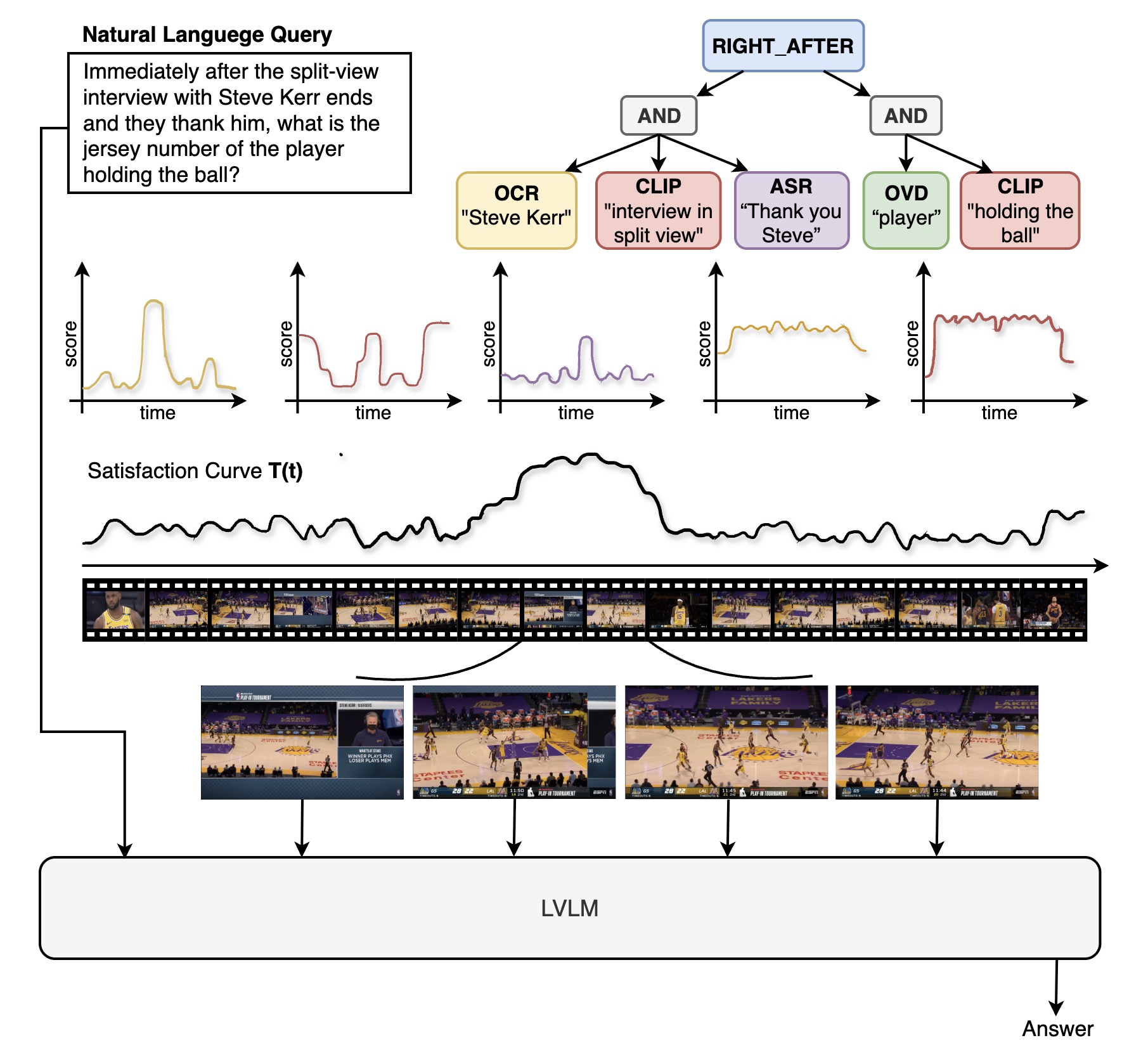}
  \caption{\textbf{HiMu pipeline overview.} Given a natural-language query, an LLM decomposes it into a hierarchical logic tree whose leaves are routed to modality-specific experts (OCR, CLIP, ASR, OVD, CLAP). Each expert produces a per-frame relevance signal over time; these signals are composed bottom-up via fuzzy-logic operators into a satisfaction curve $T(t)$. The top-scoring frames are selected and passed to the multi-modal large language model (MLLM) for answering.}
  \label{fig:supp_pipeline}
\end{figure*}

{\setcounter{tocdepth}{1}%
\renewcommand{\baselinestretch}{0.9}\small\tableofcontents\normalsize\renewcommand{\baselinestretch}{1}}

\appendix

\section{\revised{Hierarchical} Query Decomposition Details}
\label{sec:decomposition}

\subsection{System Prompt for Tree Generation}
\label{ssec:system_prompt}

The system prompt guides the LLM to decompose each question into a hierarchical logic tree. Because different benchmarks provide different audio modalities, the prompt is adapted per benchmark by enabling or disabling expert descriptions and their associated rules. We first present the \emph{common} prompt shared across all benchmarks (Listing~\ref{lst:system_prompt}), then summarize the per-benchmark adaptations in Table~\ref{tab:prompt_diff}.

\begin{lstlisting}[caption={Common system prompt for logic tree generation (shared across all benchmarks). Sections marked \texttt{[IF\_ASR]} and \texttt{[IF\_CLAP]} are included only when the corresponding expert is enabled (see Table~\ref{tab:prompt_diff}).},label={lst:system_prompt},basicstyle=\scriptsize\ttfamily,escapeinside={(*@}{@*)}]
You are a Neuro-Symbolic Logic Parser for a Multimodal
Video Understanding system. Convert Video QA questions into
structured logical trees that leverage ALL relevant modalities.

### THE EXPERTS

**Visual Experts:**
1. **OVD** - Open-Vocabulary Object Detection (YOLO-World)
   - For: Physical objects, people, and visual attributes
   - Examples: "person", "car", "dog", "red car", "man in suit"
   - Supports attribute+noun phrases. Never include numbers,
     counts, or proper names.
   - OVD CANNOT detect: countries, teams, names, states,
     emotions, actions, text, numbers

2. **OCR** - On-Screen Text Recognition
   - For: Text visible on screen - signs, labels, jersey
     numbers, names, scoreboards
   - Examples: "Exit", "10", "Warning", "Korea"

3. **CLIP** - Semantic Visual Understanding
   - For: Actions, scenes, visual states, atmosphere, abstract
     visual concepts
   - CLIP is VISUAL ONLY - queries must describe something you
     can SEE in a video frame
   - Good: "person speaking", "diagram on screen", "cooking"
   - Bad: "quantum entanglement", "innovation"
     (invisible concepts)

[IF_ASR]
4. **ASR** - Speech Recognition
   - For: Spoken words, dialogue, narration, verbal references
   - Use SHORT keywords (1-3 words), never full sentences
   - CRITICAL: People TALK about what is shown. Add an ASR
     leaf with related spoken keywords alongside visual leaves.
[/IF_ASR]

[IF_CLAP]
5. **CLAP** - Environmental Audio Events
   - For: Non-speech sounds, music, sound effects, ambient audio
   - Examples: "doorbell ringing", "applause", "glass breaking"
[/IF_CLAP]

### THE OPERATORS

- **AND**: All children must co-occur (same frame).
- **OR**: At least one child satisfied.
- **SEQ**: Temporal sequence (earliest first, latest last).
  ONLY when order is EXPLICITLY STATED.
  IMPORTANT: If the question ASKS about order/sequence,
  do NOT use SEQ. Instead: build one AND per shot,
  wrap in a flat OR. Each shot appears exactly once.
- **RIGHT_AFTER**: Immediate temporal succession.
  Exactly 2 children [cause, effect].

### KEY RULES

(*@\textcolor{revblue}{1. MULTIMODAL: Use multiple expert types when distinct observable}@*)
(*@\textcolor{revblue}{   cues support them. Never add redundant leaves only to force modality}@*)
(*@\textcolor{revblue}{   diversity. [IF\_ASR: Combine visual and audio evidence when both are}@*)
(*@\textcolor{revblue}{   relevant.] [ELSE: Combine visual experts when complementary.] }@*)
[IF_ASR]
2. ASR OVERLAP: Add ASR leaves with short keywords alongside
   visual leaves - narrators often describe what is shown.
[/IF_ASR]
3. MCQ STRUCTURE: AND(shared_context, OR(opt_1, opt_2, ...))
   - factor shared elements OUT of the OR.
4. DECOMPOSE RICH DESCRIPTIONS: Create separate leaves for
   each element: OVD for objects/people, CLIP for
   settings/states.
5. SEQ ONLY FOR KNOWN ORDER.
6. NAMES -> OCR [IF_ASR: + ASR (spoken)].
7. VISUAL STATES -> CLIP, not ASR alone.
8. META-OPTIONS: "Same", "All of the above", etc. ->
   ALWAYS skip in the OR.
9. ACTIONS IN OPTIONS: AND(OVD:object, CLIP:action).
10. TEMPORAL CAUSE: cause child = action (CLIP), not person.
11. OVERLAPPING EXPERTS encouraged.
[IF_ASR]
12. VISUAL GROUNDING: Never build ASR-only options.
13. OVERLAPPING PREDICATES: Mix experts with overlapping
    terms for robust detection.
14. SUBTITLES -> ASR + OCR.
[/IF_ASR]

### OUTPUT FORMAT
Return a single JSON object:
{"op": "AND"|"OR"|"SEQ"|"RIGHT_AFTER"|"LEAF",
 "children": [...],
 "expert": <available experts>, "query": "string"}
\end{lstlisting}

\begin{table}[tbp]
    \centering
    \caption{Per-benchmark prompt adaptations for HiMu's logic-tree parser. The common prompt (Listing~\ref{lst:system_prompt}) is shared across all benchmarks; only the dimensions shown below differ per benchmark.}
    \label{tab:prompt_diff}
    \setlength{\tabcolsep}{5pt}
    \resizebox{\columnwidth}{!}{%
    \begin{tabular}{l ccc}
        \toprule
        & \textbf{Video-MME} & \textbf{LongVideoBench} & \textbf{HERBench-Lite} \\
        \midrule
        \textbf{Active experts} & OVD, OCR, CLIP, & OVD, OCR, CLIP, & OVD, OCR, CLIP \\
        & ASR, CLAP & ASR & \\
        \midrule
        \texttt{[IF\_ASR]} included & \cmark & \cmark & \xmark \\
        \texttt{[IF\_CLAP]} included & \cmark & \xmark & \xmark \\
        \midrule
        Rules 2, 12--14 & \cmark & \cmark & \xmark \\
        \midrule
        \textbf{Prompt examples} & 4 (audio, OCR+ASR, & 3 (speech, scientific, & 3 (MCQ, ordering, \\
        & multimodal, temporal) & temporal) & OCR+visual) \\
        \bottomrule
    \end{tabular}}
\end{table}

The prompt also includes three to four worked examples tailored to each benchmark's modality regime (e.g., CLAP-based examples for Video-MME, speech-triggered examples for LongVideoBench, visual-only examples for HERBench-Lite). The complete per-benchmark prompts including all examples will be released.

\subsection{JSON Schema Constraint}
\label{ssec:json_schema}

The LLM output is constrained to valid JSON matching the following schema. Leaf nodes carry \texttt{expert} and \texttt{query} fields; internal nodes carry an \texttt{op} and a \texttt{children} array.

\begin{lstlisting}[caption={JSON schema for logic tree output.},label={lst:json_schema},basicstyle=\scriptsize\ttfamily]
{
  "op": "AND" | "OR" | "SEQ" | "RIGHT_AFTER" | "LEAF",
  "children": [<recursive tree nodes>],  // non-LEAF only
  "expert": "CLIP"|"OVD"|"OCR"|"ASR"|"CLAP",  // LEAF only
  "query": "<atomic predicate string>"          // LEAF only
}
\end{lstlisting}

The available values for \texttt{expert} are restricted per benchmark to match the active expert set (Table~\ref{tab:prompt_diff}).

\subsection{Expert Routing Rules}
\label{ssec:routing_rules}

The system prompt encodes the following routing logic, which the LLM applies when assigning predicates to experts:

\begin{itemize}[leftmargin=*,nosep]
    \item \textbf{Physical objects and people} $\to$ OVD (e.g., ``red car'', ``man in suit'').
    \item \textbf{Actions, scenes, visual states} $\to$ CLIP (e.g., ``person running'', ``sunset'').
    \item \textbf{On-screen text} $\to$ OCR (e.g., ``Exit'', ``Warning'', jersey numbers).
    \item \textbf{Spoken content and dialogue} $\to$ ASR (e.g., ``reaction'', ``careful'').
    \item \textbf{Environmental sounds} $\to$ CLAP (e.g., ``applause'', ``glass breaking'').
    \item \textbf{Proper names} $\to$ OCR + ASR (text on screen and spoken references).
    \item \textbf{Actions with objects} $\to$ AND(OVD:object, CLIP:action) for robust detection.
\end{itemize}

\noindent\textbf{Worked example.} For the question \emph{``After the doorbell rings, who opens the door?''} with options [``A man'', ``A woman''], the parser produces a tree that grounds the doorbell with the audio expert, grounds the door-opening event visually, and keeps the multiple-choice hypotheses under an \texttt{OR}:
\begin{lstlisting}[basicstyle=\scriptsize\ttfamily]
{
  "op": "RIGHT_AFTER",
  "children": [
    {
      "op": "LEAF",
      "expert": "CLAP",
      "query": "doorbell ringing"
    },
    {
      "op": "AND",
      "children": [
        {
          "op": "LEAF",
          "expert": "CLIP",
          "query": "person opens the door"
        },
        {
          "op": "OR",
          "children": [
            {
              "op": "LEAF",
              "expert": "OVD",
              "query": "man"
            },
            {
              "op": "LEAF",
              "expert": "OVD",
              "query": "woman"
            }
          ]
        }
      ]
    }
  ]
}
\end{lstlisting}
The \texttt{RIGHT\_AFTER} root captures the local causal relation expressed by the doorbell cue: high-scoring frames are those in which the visual door-opening event follows closely after the ring. The effect branch is an \texttt{AND}, requiring both the door-opening action and one of the candidate people, while the nested \texttt{OR} represents the answer alternatives.

\revon
\section{Reliability of the Logic Tree Parser}
\label{sec:parser_reliability}

HiMu evaluates parser reliability at three levels: mechanical validity, human-judged semantic fidelity, and downstream sensitivity to controlled tree corruptions. \revised{Section~\ref{subsec:ablation}} (Q5) reports the headline numbers of this evaluation; this section and Appendix~\ref{sec:tree_human_eval} give the full protocol, the per-benchmark and per-node breakdowns, and the error taxonomy behind them. This section reports mechanical validity; Appendix~\ref{sec:tree_human_eval} presents the semantic and downstream analyses. The parser-swap experiment in Appendix~\ref{ssec:llm_parser} is complementary: it measures end-task stability under different generators, not agreement with a reference interpretation.

\paragraph{Audit scope.} We audit every tree generated across Video-MME, LongVideoBench$_\text{val}$, and HERBench-Lite. A tree is counted as valid only if it parses as JSON, conforms to the schema in Appendix~\ref{ssec:json_schema}, uses only operators from the closed vocabulary, assigns every leaf both an expert and a query string, gives every internal node at least one child, respects the arity constraint on \textsc{RightAfter}, and calls no expert outside the benchmark's modality regime. These criteria are the acceptance test applied at run time: an attempt that violates any of them is rejected and regenerated, so every tree that reaches expert scoring satisfies the schema, the operator vocabulary, the node-field rules, and the benchmark's modality regime by construction. Rule violations are therefore reported over generation attempts rather than over accepted trees.

\paragraph{Retry dynamics.} Table~\ref{tab:parser_reliability} reports first-pass and post-retry success. Greedy attempt-0 decoding yields a valid tree for $98.7\%$ of questions on average, and the bounded retry budget raises this to $99.2\%$, leaving $0.8\%$ that exhaust all attempts and use a flat fallback. HERBench-Lite accounts for the observed fallbacks ($2.4\%$ on that benchmark); Video-MME and LongVideoBench$_\text{val}$ recover all first-pass failures within the retry budget.

\paragraph{Rule violations and error taxonomy.} Among failed generation attempts, $78.7\%$ are malformed JSON, $10.6\%$ contain an invalid operator value, $4.3\%$ omit a leaf field, $4.3\%$ omit internal-node children, and $2.1\%$ violate \textsc{RightAfter} arity. Rule violations on otherwise parseable output thus account for $21.3\%$ of failed attempts, and no parseable attempt invokes an expert outside its benchmark's modality regime. The bounded regeneration step handles these mechanical failures before expert scoring; semantic fidelity is evaluated separately below.

\paragraph{Operator usage.} Under the default Qwen3-VL-8B parser, \textsc{And} appears in $100.0\%$ of trees and \textsc{Or} in $94.3\%$, reflecting the shared-context and answer-option pattern. \textsc{Seq} and \textsc{RightAfter} appear in $6.8\%$ and $5.1\%$ of trees, respectively.
Across the evaluated benchmarks, mean tree depth is $2.71{\pm}0.44$. Rescaling each input signal to $[0.5,1]$ before an operator limits product attenuation in the observed shallow trees, although repeated product conjunction remains a limitation for deeper trees.
All parser and answerer calls use sampling-disabled decoding (\texttt{do\_sample=False} for local models and temperature $0$ for API models), while expert scoring, composition, and PASS are fixed under identical inputs; Random Sampling is the only stochastic condition and is reported as mean $\pm$ standard deviation over three seeded runs.

\begin{table}[tbp]
    \centering
    \revon
    \caption{Mechanical parser audit. \emph{First pass} is greedy attempt-0 decoding; \emph{with retries} applies the bounded retry budget; \emph{fallback} questions exhaust all attempts and receive a flat default tree. Validity requires schema conformance, the closed operator vocabulary, complete node fields, and the benchmark's modality regime.}
    \label{tab:parser_reliability}
    \setlength{\tabcolsep}{5pt}
    \resizebox{\columnwidth}{!}{%
    \begin{tabular}{l cccc}
        \toprule
        \textbf{Metric (\%)} & \textbf{V-MME} & \textbf{LVB$_\text{val}$} & \textbf{HERB.-Lite} & \textbf{Mean} \\
        \midrule
        First-pass success & 100.0 & 99.6 & 96.6 & 98.7 \\
        Success w/ retries & 100.0 & 100.0 & 97.6 & 99.2 \\
        Fallback (flat tree) & 0.0 & 0.0 & 2.4 & 0.8 \\
        \bottomrule
    \end{tabular}%
    }
\end{table}

\section{Human Evaluation of Generated Logic Trees}
\label{sec:tree_human_eval}

Mechanical validity guarantees that a tree is well-formed, not that it means what the question asks. We therefore complement the audit above with a human study of semantic correctness, covering operator selection, expert routing, predicate decomposition, and the sufficiency of the closed operator vocabulary.

\subsection{Study Design and Protocol}
\label{ssec:human_eval_protocol}

We sampled $270$ question--tree pairs uniformly at random, $90$ from each of Video-MME, LongVideoBench$_\text{val}$, and HERBench-Lite. Four annotators with graduate-level computer-vision training, none of whom were involved in developing the method, carried out the annotation. Each pair received a judgement against the question text, answer options, and generated tree. The fourth annotator independently reannotated $90$ pairs, comprising $30$ from each of the first three annotators' assignments, yielding $360$ judgements in total. Where the two judgements of a doubly annotated pair differ, we adopt the more severe tier, so the reported ratings are a conservative resolution of the paired annotations. The paired ratings agree on $87/90$ trees ($96.7\%$ raw agreement). Each pair agrees on $29/30$ trees, and unweighted Cohen's $\kappa$ is $0.78$, $0.74$, and $0.84$ for the three original-annotator/fourth-annotator pairs, respectively, giving a mean pairwise $\kappa$ of $0.79$. Because the tier distribution is strongly concentrated on \emph{Correct}, we report raw agreement alongside the chance-corrected coefficients.

\subsection{Evaluation Categories}
\label{ssec:human_eval_categories}

Each tree was assigned to exactly one of three tiers:
\begin{itemize}[leftmargin=*,nosep]
    \item \textbf{Correct}: the tree reflects the query semantics, routes every atomic predicate to an appropriate modality, and selects logical and temporal operators consistent with the question.
    \item \textbf{Minor flaw}: the tree contains a redundant leaf, an over-segmented predicate, or an isolated operator/routing error that, in the annotators' judgement, does not materially change the intended retrieval logic.
    \item \textbf{Major flaw}: the tree contains an incorrect operator or expert assignment that, in the annotators' judgement, materially changes the intended retrieval logic.
\end{itemize}
Annotators additionally recorded, per question, whether each operator and each expert routing decision was correct, and whether the query required a relation the four-operator vocabulary cannot express. This last judgement was collected against a named checklist of candidate relations, covering negation or exclusion, counting, ordinals, bounded intervals, and absolute temporal position, with a free-text field for any relation outside the checklist.

\subsection{Results}
\label{ssec:human_eval_results}

Table~\ref{tab:human_eval_detail} reports the benchmark-level ratings. The corresponding correct/minor/major counts are $84/5/1$ for Video-MME, $81/8/1$ for LongVideoBench$_\text{val}$, and $84/3/3$ for HERBench-Lite. Fully correct ratings range from $90.0\%$ to $93.3\%$, while major flaws range from $1.1\%$ to $3.3\%$. Pooled over the $270$ trees, the tier counts are $249/16/5$, or $92.2/5.9/1.9\%$. Operator accuracy ranges from $90.9\%$ to $100\%$ across operator nodes, and expert-routing accuracy ranges from $90.9\%$ to $100\%$ across expert-routed leaves. Two remarks on how to read these node-level rates. First, they are computed over nodes rather than trees, so a benchmark with a high per-node accuracy can still contain a few flawed trees, and the tree-level tiers above are the stricter summary. Second, the entries below $100\%$ are concentrated in the \textsc{Seq} nodes of LongVideoBench$_\text{val}$, the \textsc{Or} nodes of HERBench-Lite, and the OCR, OVD, and ASR leaves, whereas \textsc{And} and CLIP routing are correct on every judged node in all three benchmarks.

\begin{table}[tbp]
    \centering
    \revon
    \caption{Human evaluation of generated logic trees ($N{=}270$ unique pairs; $90$ per benchmark; $90$ pairs independently reannotated with $96.7\%$ raw agreement and mean pairwise Cohen's $\kappa{=}0.79$). Overall quality is measured per tree, whereas operator and routing accuracies are measured over the corresponding nodes. ``N/A'' marks experts unavailable under a benchmark's modality regime.}
    \label{tab:human_eval_detail}
    \setlength{\tabcolsep}{5pt}
    \resizebox{\columnwidth}{!}{%
    \begin{tabular}{llccc}
        \toprule
        \textbf{Category} & \textbf{Sub-metric} & \textbf{V-MME} & \textbf{LVB$_\text{val}$} & \textbf{HERB.-Lite} \\
        \midrule
        \multirow{3}{*}{\textbf{Overall quality}}
         & Correct (\%) & 93.3 & 90.0 & 93.3 \\
         & Minor flaw (\%) & 5.6 & 8.9 & 3.3 \\
         & Major flaw (\%) & 1.1 & 1.1 & 3.3 \\
        \midrule
        \multirow{4}{*}{\textbf{Operator accuracy}}
         & \textsc{And} (\%) & 100.0 & 100.0 & 100.0 \\
         & \textsc{Or} (\%) & 100.0 & 100.0 & 99.1 \\
         & \textsc{Seq} (\%) & 100.0 & 90.9 & 100.0 \\
         & \textsc{RightAfter} (\%) & 100.0 & 100.0 & 100.0 \\
        \midrule
        \multirow{5}{*}{\textbf{Expert routing}}
         & CLIP (\%) & 100.0 & 100.0 & 100.0 \\
         & CLAP (\%) & 100.0 & N/A & N/A \\
         & ASR (\%) & 93.8 & 97.1 & N/A \\
         & OCR (\%) & 94.1 & 90.9 & 100.0 \\
         & OVD (\%) & 94.7 & 97.9 & 97.3 \\
        \midrule
        \textbf{Operator coverage} & Missing operator (\%) & 3.3 & 1.1 & 0.0 \\
        \bottomrule
    \end{tabular}%
    }
\end{table}

\subsection{Minor Flaws and Out-of-Vocabulary Relations}
\label{ssec:human_eval_analysis}

\paragraph{Predicate over-segmentation.} The predominant minor-flaw mode is an over-segmented visual description. For the query \emph{``a man with light hair wearing a beige jacket is running''}, for example, the parser may emit \textsc{clip}(\texttt{``a man with light hair wearing a beige jacket''}) and \textsc{clip}(\texttt{``man is running''}) under an \textsc{And} rather than one combined predicate. This preserves the intended expert and conjunction but introduces a redundant factor, matching the definition of a minor rather than major flaw.

\paragraph{Operator vocabulary coverage.} Annotators judged the four operators sufficient for $100\%$ of sampled HERBench-Lite queries, $98.9\%$ of LongVideoBench$_\text{val}$, and $96.7\%$ of Video-MME (Table~\ref{tab:human_eval_detail}). Every flagged pair requests an absolute temporal anchor (\eg \emph{``at the end of the video''}), for which HiMu drops the unsupported positional constraint and matches the remaining predicate over the full sequence; negation or exclusion, counting, ordinals, and bounded intervals are never flagged. This asymmetry follows from what selection is asked to do: HiMu retrieves the evidence a question refers to, while the answerer computes the answer from it, so a relation that constrains the answer rather than the evidence is satisfied by covering the candidate evidence. For \emph{``which object never appears?''}, the option branches under \textsc{Or} cover each candidate and the intended answer is the one absent from every selected frame; for \emph{``how many times does the dog enter the house?''}, \textsc{And}(\textsc{clip}(\texttt{``dog enters the house''}), \textsc{ovd}(\texttt{``dog''})) scores exactly the moments to be counted, which PASS (Appendix~\ref{sec:pass}) spreads across the distinct peaks of $T(t)$. An absolute anchor instead restricts which frames are relevant, so evidence coverage cannot substitute for it. More generally, the vocabulary has no explicit operator for negation or exclusion, counting, ordinals, bounded intervals, or absolute position; such relations may be retained as natural language inside a leaf query but are not enforced during composition, so the coverage rates characterize these samples rather than universal expressivity.

\paragraph{Downstream sensitivity.} The controlled structural ablations in \revised{Table~\ref{tab:expert_ablation}} perturb the tree while holding its leaves, expert outputs, answerer, and frame budget fixed. Substituting one operator type costs $1.0$--$2.5$\,pp, removing nesting costs $1.4$\,pp, and replacing the tree with flat fusion costs $5.5$\,pp. These interventions are stress tests rather than estimates of the frequency of naturally occurring parser errors; they show that operator and hierarchy errors can materially affect the selected evidence, motivating the direct quality audit above.
\revised{A paired correction study directly measures the naturally occurring errors: we manually correct the 21 trees labeled with minor or major flaws and rerun the standard HiMu pipeline with Qwen3-VL-8B, holding the answerer, frame budget, and expert configuration fixed so that the corrected tree is the only difference. Predictions are identical for 19/21 questions. Both changes involve major-flaw trees: the HERBench-Lite prediction changes from incorrect to correct, whereas the LongVideoBench$_\text{val}$ prediction changes between two incorrect answers, yielding one net correction and no correct-to-incorrect transition. Accuracy on this 21-question subset therefore rises by $4.8$\,pp. Two properties follow. First, every minor flaw is prediction-preserving here, consistent with the tier definition in Appendix~\ref{ssec:human_eval_categories}: over-segmented predicates add redundant conjuncts that leave the frame ranking essentially unchanged. Second, the effect of the residual parser errors is an order of magnitude smaller than the $1.4$--$5.5$\,pp cost of the deliberate corruptions above, because those corruptions perturb every tree in the evaluation set whereas semantic flaws occur in $7.8\%$ of trees and materially alter retrieval logic in only $1.9\%$. Compositional structure is thus both the main source of HiMu's gains and a component the parser supplies reliably at the scale of the full benchmarks.}
\revoff

\section{Implementation Details}
\label{sec:implementation}

Table~\ref{tab:hyperparams} lists all hyperparameter values used throughout the paper. Parameters are grouped by the pipeline stage they belong to, with references to the corresponding equations and sections \revised{of Section~\ref{sec:method}}.

\begin{table*}[tbp]
    \centering
    \caption{Complete hyperparameter settings for HiMu.}
    \label{tab:hyperparams}
    \setlength{\tabcolsep}{5pt}
    \begin{tabular}{l l l p{5.5cm}}
        \toprule
        \textbf{Parameter} & \textbf{Symbol} & \textbf{Value} & \textbf{Description} \\
        \midrule
        \multicolumn{4}{l}{\textit{Normalization (\revised{Eq.~\ref{eq:norm}, Sec.~\ref{ssec:experts}})}} \\
        Sigmoid sharpness & $\gamma$ & 3.0 & Controls the steepness of the sigmoid in median/MAD normalization; higher values yield sharper contrast between high- and low-scoring frames \\
        MAD stabilizer & $\delta$ & $10^{-6}$ & Prevents division by zero when all expert scores are identical (MAD\,$=$\,0) \\
        \midrule
        \multicolumn{4}{l}{\textit{Temporal operators (\revised{Eqs.~\ref{eq:ra_eff}--\ref{eq:ra}, Sec.~\ref{ssec:composition}})}} \\
        RightAfter decay & $\kappa$ & 2.0 & Exponential decay rate for the \textsc{RightAfter} operator; larger values restrict causal pairing to temporally closer frames \\
        \midrule
        \multicolumn{4}{l}{\textit{Bandwidth-matched smoothing (\revised{Eq.~\ref{eq:smooth}, Sec.~\ref{ssec:experts}})}} \\
        CLIP smoothing & $\sigma_\text{clip}$ & 0.5 & Narrow kernel for frame-precise visual similarity scores \\
        OVD smoothing & $\sigma_\text{ovd}$ & 0.5 & Narrow kernel for frame-precise object detection confidences \\
        OCR smoothing & $\sigma_\text{ocr}$ & 0.5 & Narrow kernel for frame-precise on-screen text detections \\
        ASR smoothing & $\sigma_\text{asr}$ & 1.5 & Wider kernel to bridge the coarser temporal resolution of speech transcripts \\
        CLAP smoothing & $\sigma_\text{clap}$ & 2.0 & Widest kernel to account for the coarse temporal granularity of environmental audio events \\
        \midrule
        \multicolumn{4}{l}{\textit{PASS selection (\revised{Sec.~\ref{ssec:selection}})}} \\
        Number of peaks & $N_p$ & $\lfloor\sqrt{K}\rfloor$ & Number of peaks to detect; scales sub-linearly with $K$ to encourage coverage of multiple relevant events rather than concentrating on a single segment \\
        Neighbors per peak & $N_n$ & $\lfloor\sqrt{K}/2\rfloor$ & Neighbors added around each peak to capture fine-grained details, small movements, and short-term temporal changes within each key moment \\
        Local window size & $w$ & $\lfloor\sqrt{K}\rfloor$ & Total width of the neighbor search window centered at each peak ($\lfloor w/2\rfloor$ frames per side); equal to $\Delta$ to prevent overlap between adjacent peak neighborhoods \\
        Min inter-peak distance & $\Delta$ & $\lfloor\sqrt{K}\rfloor$ & Minimum frame separation between peaks; uses the same formula as $N_p$ ($\lfloor\sqrt{K}\rfloor$); this equality ensures non-overlapping temporal neighborhoods and diverse event coverage \\
        \bottomrule
    \end{tabular}
\end{table*}

\section{PASS Algorithm}
\label{sec:pass}

\begin{figure*}[htbp]
    \centering
    \includegraphics[width=\linewidth]{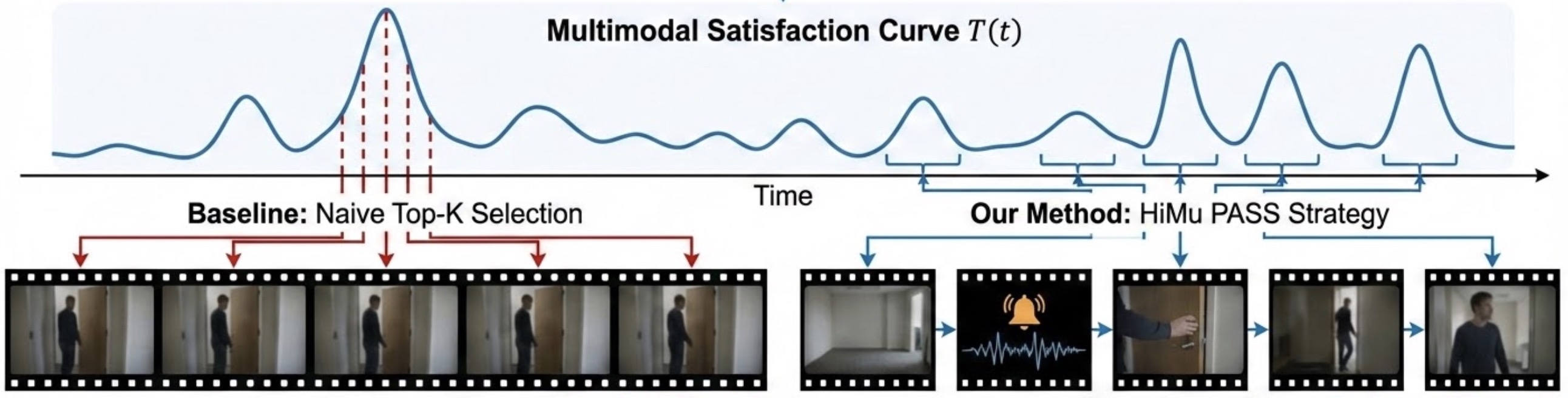}
    \caption{\textbf{PASS vs.\ naive top-$K$ selection.} Given the multimodal satisfaction curve $T(t)$, naive top-$K$ (left, red) concentrates all selected frames around the single highest peak, missing other relevant events. PASS (right, blue) detects multiple peaks, spreads neighbors around each, and fills the remaining budget greedily, yielding diverse temporal coverage.}
    \label{fig:pass_illustration}
\end{figure*}

Algorithm~\ref{alg:pass} provides the pseudocode for PASS (Peak-And-Spread Selection). Phase~1 identifies $N_p$ prominent peaks in the satisfaction curve while enforcing a minimum temporal separation of $\Delta$ frames, preventing redundant selection from a single high-scoring segment without artificially requiring coverage of the full timeline. Phase~2 augments each peak with its $N_n$ highest-scoring neighbors within a local window of $w$ frames, capturing short-term motion context. Phase~3 fills the remaining budget greedily from the highest-scoring unselected frames.

\paragraph{PASS vs.\ vanilla top-$K$.} As illustrated in Fig.~\ref{fig:pass_illustration}, on the full Video-MME test set (Qwen3-VL-8B, $K{=}16$), PASS achieves 73.23\% compared to 72.37\% for vanilla top-$K$ selection ($+$0.86\,pp). Vanilla top-$K$ tends to concentrate all selected frames within a single high-scoring segment, missing relevant events elsewhere in the video. PASS mitigates this by explicitly detecting multiple peaks and spreading neighbors around each, ensuring coverage of distinct temporal events while preserving fine-grained context around each key moment.

\revised{\paragraph{Random-sampling control.} We additionally sample $K{=}16$ distinct frames uniformly at random from the same 1\,fps candidate set used by the other selectors. We run the complete evaluation with seeds 41, 42, and 43 and report the mean $\pm$ standard deviation across the three runs. The resulting accuracies are $74.7{\pm}0.2$, $64.7{\pm}0.3$, $53.6{\pm}0.3$, and $64.4{\pm}0.3$ on the Video-MME Short, Medium, Long, and Overall splits, respectively; $55.8{\pm}0.3$ on LongVideoBench$_\text{val}$; and $41.4{\pm}0.1$ on HERBench-Lite.}

\begin{algorithm}[t]
\caption{PASS: Peak-And-Spread Selection}
\label{alg:pass}
\begin{algorithmic}[1]
\REQUIRE Satisfaction curve $T(t)$ for $t \in \{1,\dots,N\}$, frame budget $K$, number of peaks $N_p$, neighbors per peak $N_n$, window size $w$, min inter-peak distance $\Delta$
\ENSURE Selected frame indices $\mathcal{S}$ with $|\mathcal{S}| = K$
\STATE $\mathcal{S} \leftarrow \emptyset$
\STATE
\STATE \textbf{// Phase 1: Peak detection}
\STATE $\mathcal{P} \leftarrow \emptyset$ \COMMENT{set of selected peak indices}
\STATE $\mathcal{C} \leftarrow \{t : T(t) > T(t{-}1) \;\text{and}\; T(t) > T(t{+}1)\}$ \COMMENT{find all local maxima in the satisfaction curve}
\STATE Sort $\mathcal{C}$ by $T(t)$ in descending order \COMMENT{process highest-scoring peaks first}
\FOR{$t \in \mathcal{C}$}
    \IF{$|t - t'| \geq \Delta$ for all $t' \in \mathcal{P}$}
        \STATE $\mathcal{P} \leftarrow \mathcal{P} \cup \{t\}$ \COMMENT{enforce min separation: skip peaks too close to existing ones}
    \ENDIF
    \IF{$|\mathcal{P}| = N_p$}
        \STATE \textbf{break} \COMMENT{enough diverse peaks collected}
    \ENDIF
\ENDFOR
\STATE $\mathcal{S} \leftarrow \mathcal{S} \cup \mathcal{P}$
\STATE
\STATE \textbf{// Phase 2: Neighbor spread} \COMMENT{add nearby frames around each peak to capture local temporal context}
\FOR{each peak $p \in \mathcal{P}$}
    \STATE $\mathcal{W}_p \leftarrow \{t : |t - p| \leq \lfloor w/2 \rfloor,\; t \notin \mathcal{S}\}$ \COMMENT{candidate neighbors within window $w$ around peak $p$}
    \STATE Sort $\mathcal{W}_p$ by $T(t)$ in descending order
    \STATE $\mathcal{S} \leftarrow \mathcal{S} \cup \mathcal{W}_p[1:N_n]$ \COMMENT{keep top-$N_n$: best context frames around this peak}
\ENDFOR
\STATE
\STATE \textbf{// Phase 3: Greedy fill} \COMMENT{use remaining budget on highest-scoring uncovered frames}
\STATE $\mathcal{R} \leftarrow \{1,\dots,N\} \setminus \mathcal{S}$ \COMMENT{frames not yet selected}
\STATE Sort $\mathcal{R}$ by $T(t)$ in descending order
\STATE $\mathcal{S} \leftarrow \mathcal{S} \cup \mathcal{R}[1:(K - |\mathcal{S}|)]$ \COMMENT{fill to budget $K$}
\RETURN $\mathcal{S}$
\end{algorithmic}
\end{algorithm}

\section{Sensitivity Analysis}
\label{sec:sensitivity}

Unless otherwise noted, sensitivity experiments use Qwen3-VL-8B as the downstream MLLM with $K{=}16$ frames on the full Video-MME test set. Each experiment changes exactly one component from the default configuration, isolating the contribution of each design choice.

\subsection{Hyperparameter Sensitivity}
\label{ssec:hyperparam_sensitivity}

We systematically vary the smoothing bandwidths, temporal decay factor, and sigmoid sharpness one at a time (Table~\ref{tab:hyperparam_sensitivity}). All perturbations remain within at most 0.71\,pp of the baseline, confirming that HiMu is not tightly coupled to any single hyperparameter setting. This stability stems from the \revised{compositional architecture: frame relevance is determined by the structure of the logic tree} rather than by raw expert scores, so moderate changes to score processing have limited effect on the final frame ranking.

\noindent\textbf{Smoothing bandwidths.}
Smoothing has a small overall effect: disabling it entirely ($+$0.49\,pp) or zeroing visual smoothing ($+$0.26\,pp) leaves accuracy essentially unchanged, while over-smoothing the visual signal (visual $\sigma{=}2$) is the most harmful setting ($-$0.71\,pp), blurring shot boundaries. For speech, removing smoothing has no effect (0.00\,pp) and over-smoothing ($\sigma_\text{asr}{=}4$) costs only $-$0.41\,pp, confirming that the default $\sigma_\text{asr}{=}1.5$ is a safe choice.

\noindent\textbf{Temporal decay $\kappa$.}
Across a wide range of decay rates ($\kappa \in \{0.5, 1.0, 4.0, 8.0\}$), accuracy varies by at most $-$0.08\,pp, showing that the \textsc{RightAfter} operator is highly robust to the exact decay setting.

\noindent\textbf{Sigmoid sharpness $\gamma$.}
Reducing the sigmoid contrast ($\gamma{=}1.0$) slightly improves accuracy ($+$0.56\,pp), while increasing it ($\gamma{=}5.0$) costs $-$0.23\,pp; both remain within 0.56\,pp of the default, demonstrating that the median/MAD normalization is robust to the exact sharpness setting.

\begin{table}[tbp]
    \centering
    \caption{Hyperparameter sensitivity on the full Video-MME test set (Qwen3-VL-8B, $K{=}16$). Each row changes one hyperparameter from the default. $\Delta$ is the difference from the baseline.}
    \label{tab:hyperparam_sensitivity}
    \setlength{\tabcolsep}{8pt}
    \resizebox{\columnwidth}{!}{%
    \begin{tabular}{l c c}
        \toprule
        \textbf{Configuration} & \textbf{Accuracy} & \textbf{$\Delta$} \\
        \midrule
        \rowcolor{gray!10} \textbf{HiMu (default)} & \textbf{73.23} & 0.00 \\
        \midrule
        \multicolumn{3}{l}{\textit{Smoothing bandwidths}} \\
        All smoothing disabled & 73.72 & $+$0.49 \\
        Visual $\sigma$ (CLIP/OVD/OCR, default: 0.5) $\to$ 0 & 73.49 & $+$0.26 \\
        Visual $\sigma$ (CLIP/OVD/OCR, default: 0.5) $\to$ 2 & 72.52 & $-$0.71 \\
        Speech $\sigma$ (ASR, default: 1.5) $\to$ 0 & 73.23 & 0.00 \\
        Speech $\sigma$ (ASR, default: 1.5) $\to$ 4 & 72.82 & $-$0.41 \\
        \midrule
        \multicolumn{3}{l}{\textit{Temporal decay factor $\kappa$ (default: 2.0)}} \\
        $\kappa{=}0.5$ & 73.19 & $-$0.04 \\
        $\kappa{=}1.0$ & 73.19 & $-$0.04 \\
        $\kappa{=}4.0$ & 73.15 & $-$0.08 \\
        $\kappa{=}8.0$ & 73.15 & $-$0.08 \\
        \midrule
        \multicolumn{3}{l}{\textit{Sigmoid sharpness $\gamma$ (default: 3.0)}} \\
        $\gamma{=}1.0$ & 73.79 & $+$0.56 \\
        $\gamma{=}5.0$ & 73.00 & $-$0.23 \\
        \bottomrule
    \end{tabular}%
    }
\end{table}

\revon
\subsection{PASS Parameter Sensitivity}
\label{ssec:pass_sensitivity}

PASS exposes four quantities: the number of peaks $N_p$, the neighbors added per peak $N_n$, the local window width $w$, and the minimum inter-peak distance, written here as $\Delta_p$ to avoid confusion with the accuracy difference column. All four are derived from the frame budget by a single rule rather than tuned ($N_p = w = \Delta_p = \lfloor\sqrt{K}\rfloor$ and $N_n = \lfloor\sqrt{K}/2\rfloor$, giving $4, 4, 4, 2$ at $K{=}16$). Table~\ref{tab:pass_sensitivity} varies each one independently over a $2\times$ range in both directions while holding the others at their derived values.

All eight perturbations remain within $0.72$\,pp of the default, and six remain within $0.25$\,pp. Halving or doubling $N_p$ changes accuracy by $+0.07$ and $+0.23$\,pp; the Phase~3 greedy fill limits sensitivity to the number of detected peaks. Increasing $N_n$ from $2$ to $4$ changes accuracy by $-0.36$\,pp, consistent with allocating more of the fixed budget near existing peaks. Varying $\Delta_p$ changes accuracy by at most $0.24$\,pp. The largest change is $-0.72$\,pp at $w{=}2$, while $w{=}8$ matches the default.
For comparison, replacing PASS with vanilla top-$K$ in Table~\ref{tab:backbone_ablation} costs $0.86$\,pp, more than any one-parameter perturbation. The peak-and-spread rule is therefore locally robust around the budget-derived defaults and improves over the simpler selection rule.

\begin{table}[tbp]
    \centering
    \revon
    \caption{PASS parameter sensitivity on the full Video-MME test set (Qwen3-VL-8B, $K{=}16$). Each row changes one PASS parameter from its derived default while holding the rest fixed. $\Delta$ is the accuracy difference from the baseline; $\Delta_p$ denotes the minimum inter-peak distance.}
    \label{tab:pass_sensitivity}
    \setlength{\tabcolsep}{8pt}
    \resizebox{\columnwidth}{!}{%
    \begin{tabular}{l c c}
        \toprule
        \textbf{Configuration} & \textbf{Accuracy} & \textbf{$\Delta$} \\
        \midrule
        \rowcolor{gray!10} \textbf{HiMu (default: $N_p{=}4$, $N_n{=}2$, $w{=}4$, $\Delta_p{=}4$)} & \textbf{73.23} & 0.00 \\
        \midrule
        \multicolumn{3}{l}{\textit{Number of peaks $N_p$ (default: 4)}} \\
        $N_p{=}2$ & 73.30 & $+$0.07 \\
        $N_p{=}8$ & 73.46 & $+$0.23 \\
        \midrule
        \multicolumn{3}{l}{\textit{Neighbors per peak $N_n$ (default: 2)}} \\
        $N_n{=}1$ & 73.30 & $+$0.07 \\
        $N_n{=}4$ & 72.87 & $-$0.36 \\
        \midrule
        \multicolumn{3}{l}{\textit{Min inter-peak distance $\Delta_p$ (default: 4)}} \\
        $\Delta_p{=}2$ & 73.23 & 0.00 \\
        $\Delta_p{=}8$ & 72.99 & $-$0.24 \\
        \midrule
        \multicolumn{3}{l}{\textit{Local window width $w$ (default: 4)}} \\
        $w{=}2$ & 72.51 & $-$0.72 \\
        $w{=}8$ & 73.23 & 0.00 \\
        \bottomrule
    \end{tabular}%
    }
\end{table}

\revoff
\subsection{Expert Backbone Ablation}
\label{ssec:backbone_ablation}

To evaluate whether HiMu's gains depend on a specific set of pretrained models, we swap each expert's underlying backbone one at a time (Table~\ref{tab:backbone_ablation}). All substitutions stay within $0.6$\,pp of the default, with Grounding DINO ($+$0.52\,pp) and the music-speech CLAP ($+$0.07\,pp) even yielding marginal improvements. This confirms that HiMu's strength lies in the hierarchical composition of multimodal signals rather than in any single expert backbone, allowing practitioners to substitute backbones based on deployment constraints with minimal accuracy degradation.

\begin{table}[tbp]
    \centering
    \caption{Component-level ablation on the full Video-MME test set (Qwen3-VL-8B, $K{=}16$): expert backbone swaps and selection strategy. Each row changes one component from the default configuration. $\Delta$ is the difference from the baseline.}
    \label{tab:backbone_ablation}
    \setlength{\tabcolsep}{8pt}
    \resizebox{\columnwidth}{!}{%
    \begin{tabular}{l c c}
        \toprule
        \textbf{Configuration} & \textbf{Accuracy} & \textbf{$\Delta$} \\
        \midrule
        \rowcolor{gray!10} \textbf{HiMu (default)} & \textbf{73.23} & 0.00 \\
        \midrule
        \multicolumn{3}{l}{\textit{Expert backbone swap}} \\
        CLIP-dfn~\cite{dfn2024} $\rightarrow$ SigLIP2~\cite{siglip2024} & 73.15 & $-$0.08 \\
        YOLO-World v2~\cite{yoloworld2024} $\rightarrow$ Grounding DINO~\cite{groundingdino2024} & 73.75 & +0.52 \\
        docTR~\cite{doctr2021} $\rightarrow$ EasyOCR~\cite{easyocr2020} & 72.63 & $-$0.60 \\
        faster-whisper large-v3-turbo~\cite{whisper2023} $\rightarrow$ whisper-large~\cite{whisper2023} & 72.63 & $-$0.60 \\
        LAION CLAP~\cite{clap2023} $\rightarrow$ CLAP music-speech~\cite{clap2023} & 73.30 & +0.07 \\
        \midrule
        \multicolumn{3}{l}{\textit{Selection strategy}} \\
        PASS $\rightarrow$ Vanilla top-$K$ & 72.37 & $-$0.86 \\
        \bottomrule
    \end{tabular}%
    }
\end{table}

\subsection{LLM Tree Parser Comparison}
\label{ssec:llm_parser}

In the default configuration, Qwen3-VL-8B serves as both the tree parser and the downstream MLLM answerer; here we hold the answerer fixed (Qwen3-VL-8B) and vary only the LLM that generates the logic tree, evaluating on the full Video-MME test set (Table~\ref{tab:llm_parser}). All four parsers, spanning open-source and proprietary models of different scales, achieve accuracy within a 0.78\,pp range. This tight spread demonstrates that HiMu's structured prompt and JSON schema constraint (Appendix~\ref{ssec:json_schema}) effectively guide diverse LLMs toward trees that yield near-equivalent selection accuracy, making the system robust to the choice of tree parser. The proprietary Gemini-2.5-Flash is marginally best ($+$0.07\,pp), and even the weakest parser (LLaVA-OV-1.5-8B) reaches 72.52\%, only 0.71\,pp below the default.

\begin{table}[tbp]
    \centering
    \caption{Effect of LLM tree parser on accuracy over the full Video-MME test set (Qwen3-VL-8B as MLLM answerer). $\Delta$ is the difference from the Qwen3-VL baseline.}
    \label{tab:llm_parser}
    \setlength{\tabcolsep}{8pt}
    \begin{tabular}{l c c}
        \toprule
        \textbf{Tree Parser LLM} & \textbf{Accuracy} & \textbf{$\Delta$} \\
        \midrule
        \rowcolor{gray!10} Qwen3-VL-8B~\cite{qwen3vl2025} (default) & 73.23 & 0.00 \\
        \textbf{Gemini-2.5-Flash~\cite{gemini25_2025}} & \textbf{73.30} & \textbf{+0.07} \\
        InternVL-3.5-8B~\cite{internvl35_2025} & 72.82 & $-$0.41 \\
        LLaVA-OV-1.5-8B~\cite{llavaov2024} & 72.52 & $-$0.71 \\
        \bottomrule
    \end{tabular}
\end{table}

\subsection{Component Ablation Protocol}
\label{ssec:structural_ablation}

\revised{Section~\ref{subsec:ablation}} reports expert, visual-only, and structural ablations on the full Video-MME test set with Qwen3-VL-8B as the downstream MLLM and $K{=}16$ selected frames; the full-tree reference scores $73.23\%$ overall.

\paragraph{Expert leave-one-out.} The expert rows use a fixed-tree intervention: each question is parsed once with the full expert set, then all leaves of the ablated expert are neutralized before score composition. A removed leaf is replaced by the neutral constant of its parent operator: 1 under \textsc{And}, \textsc{Seq}, and \textsc{RightAfter}, and 0 under \textsc{Or}. This keeps the query decomposition, topology, remaining expert outputs, answerer, and frame budget fixed, isolating the marginal evidence carried by the removed expert.

\paragraph{Visual-only configuration.} HiMu-Visual evaluates the adaptive visual-only setting: the tree parser prompt and JSON schema expose only CLIP, OVD, and OCR, so each question is reparsed before scoring with the visual experts. This measures the deployable behavior of HiMu when speech and non-speech audio experts are unavailable.
\revised{\paragraph{Matched fixed-tree control.} Jointly neutralizing ASR and CLAP in the original multimodal tree yields $71.1\%$ ($-2.1$\,pp). With the same visual-only expert set, HiMu-Visual reaches $72.8\%$ after reparsing, quantifying a $1.7$\,pp recovery from adaptive planning.}

\paragraph{Structural ablations.} Each structural condition perturbs the parsed logic tree before score composition while keeping the expert outputs, downstream answerer, and frame budget fixed.

\paragraph{Operator substitutions.} The four operator substitutions leave the tree topology \emph{byte-identical} (same internal nodes, same children, same leaves and DFS order; leaves are never modified); only the fusion \emph{function} at each matching node changes. Per-frame signals are rescaled to $[0.5,1.0]$ before each operator, so a substitution alters how children are combined but not the underlying evidence:
\begin{itemize}[leftmargin=*,nosep]
    \item \textbf{\textsc{And}$\to$\textsc{Or}} (\textsc{no\_and}): every conjunction becomes a disjunction (probabilistic sum), giving the loosest reading, ``any child active'' instead of ``all children active''.
    \item \textbf{\textsc{Or}$\to$\textsc{And}} (\textsc{no\_or}): every disjunction becomes a conjunction (product t-norm), giving the strictest reading, ``all'' instead of ``any''.
    \item \textbf{\textsc{Seq}$\to$\textsc{And}} (\textsc{no\_seq}): temporal ordering is dropped, retaining only product co-occurrence of the same children.
    \item \textbf{\textsc{RightAfter}$\to$\textsc{And}} (\textsc{no\_right\_after}): the exponentially-decayed cause$\to$effect coupling ($\kappa{=}2$) is dropped, retaining only co-occurrence.
\end{itemize}

\paragraph{Flattened hierarchy.} The \textbf{w/o nesting} condition (\textsc{no\_nesting}) is the only one that changes topology: all leaves are collected depth-first and wrapped under a \emph{single} node that applies the tree's original \emph{root} operator. This removes hierarchy while preserving the LLM's top-level operator choice. It is therefore milder than Flat Fusion, which additionally forces a generic flat combine (\textsc{simple\_or}) rather than the query-appropriate root operator.

\revon
\section{Scope and Benchmark Suitability}
\label{sec:scope}

The main evaluation targets long-form, multi-evidence retrieval. We additionally evaluate NExT-QA~\cite{nextqa2021}, with shorter videos averaging $44$\,s, and EgoSchema~\cite{egoschema2023}, which emphasizes egocentric, video-level understanding. Both use Qwen3-VL-8B at $K{=}16$ frames, matching the main-paper protocol.

\begin{table}[tbp]
    \centering
    \revon
    \caption{Additional benchmarks (Qwen3-VL-8B, $K{=}16$). The advantage over uniform sampling narrows on shorter videos and video-level questions.}
    \label{tab:scope}
    \setlength{\tabcolsep}{8pt}
    \begin{tabular}{l c c c}
        \toprule
        \textbf{Benchmark} & \textbf{Uniform} & \textbf{HiMu} & \textbf{$\Delta$} \\
        \midrule
        NExT-QA~\cite{nextqa2021} & 81.9 & \textbf{82.7} & $+$0.8 \\
        EgoSchema~\cite{egoschema2023} & \textbf{74.0} & 73.1 & $-$0.9 \\
        \bottomrule
    \end{tabular}
\end{table}

\paragraph{Short temporal horizons.} On NExT-QA, HiMu improves over uniform sampling by $0.8$\,pp. The smaller margin follows from how much of the timeline the budget already covers: at an average clip length of $44$\,s, $K{=}16$ frames amount to one frame every $2.8$\,s, so uniform sampling alone retains roughly a third of the $1$\,fps candidate pool that HiMu scores. On the Video-MME Long split, the same budget retains under $1\%$ of that pool. When near-uniform coverage rarely misses the relevant moment, the headroom available to any selector is small, and the Video-MME Short split shows the same trend as the budget grows relative to video length.

\paragraph{Video-level questions.} On EgoSchema, HiMu is $0.9$\,pp below uniform. This result indicates that localized evidence scoring is not uniformly beneficial for questions whose answer depends on the activity or narrative across the video.

\paragraph{Operating regime.} Taken with the main results, these experiments position HiMu for videos that are long relative to the frame budget and questions that name localizable, possibly cross-modal evidence. Uniform sampling is a stronger cost--accuracy choice when the budget already provides dense coverage or the question requires a global summary.
\revoff

\section{Selector Latency Breakdown}
\label{sec:latency}

Table~\ref{tab:latency_breakdown} reports HiMu's selector-only forward-pass compute for a 10-minute video sampled at 1\,FPS (600 candidate frames, $K{=}16$) on 8$\times$ RTX 6000 Pro GPUs; OCR is evaluated at half rate on 300 sampled frames.
We measure the selector inference path with model weights loaded and prepared inputs already placed on the assigned devices.

\begin{table*}[t]
    \centering
    \caption{Per-component selector forward-pass compute and latency for HiMu on 8$\times$ RTX 6000 Pro GPUs. A 10-minute video at 1\,FPS yields 600 candidate frames; OCR uses 300 half-rate sampled frames. Preprocessing is cacheable once per video; per-query stages run for each question. We measure the selector inference path with model weights loaded and prepared inputs already placed on the assigned devices. Total rows report wall-clock critical-path latency under the parallel schedule, not the sum of all row latencies. $^\dagger$Conditional on the corresponding expert appearing in the logic tree; when absent, the component is skipped entirely.}
    \label{tab:latency_breakdown}
    \setlength{\tabcolsep}{5pt}
    \renewcommand{\arraystretch}{1.12}
    \resizebox{\textwidth}{!}{%
    \begin{tabular}{p{0.25\textwidth} p{0.36\textwidth} c r r}
        \toprule
        \textbf{Component} & \textbf{Description} & \textbf{HW} & \textbf{TFLOPs} & \textbf{Latency (s)} \\
        \midrule
        \multicolumn{5}{l}{\textit{Preprocessing (cacheable, once per video)}} \\
        CLIP-dfn ViT-L/14 & visual semantics; 600 frames @224 & 1 GPU & 96.2 & 0.75 \\
        LAION CLAP$^\dagger$ & audio events; 300 win @2\,s & 1 GPU & 3.5 & 0.20 \\
        docTR OCR\newline (db\_resnet50+parseq)$^\dagger$ & on-screen text; 300 frames (${\sim}$2.5k crops) & 2 GPUs & 46.3 & 1.72 \\
        faster-whisper lv3-turbo$^\dagger$ (est.) & speech; full 600\,s audio & 4 GPUs & 11.5 & 2.74 \\
        \rowcolor{gray!10} \textbf{Preprocessing total} & busiest GPU & 8 GPUs & \textbf{96.2} & \textbf{2.74} \\
        \midrule
        \multicolumn{5}{l}{\textit{Per-query (uncacheable)}} \\
        Qwen3-VL-8B parser & query$\to$logic tree; 2348+67 tok & 1 GPU & 37.0 & 1.58 \\
        YOLOv8x-worldv2$^\dagger$ & open-vocab detection; 600 frames & 6 GPUs & 16.1 & 0.31 \\
        Per-query scoring & CLIP text+cos., ASR match; 1 query & 1 GPU & ${<}$0.05 & 0.20 \\
        Composition + PASS & normalize/smooth/logic/select & CPU & ${<}$0.05 & ${<}$0.01 \\
        \rowcolor{gray!10} \textbf{Per-query total} & & & \textbf{53.1} & \textbf{1.88} \\
        \midrule
        \rowcolor{gray!10} \textbf{First-query total} & & & \textbf{149.4} & \textbf{4.63} \\
        \bottomrule
    \end{tabular}%
    }
\end{table*}

\paragraph{Preprocessing.}
Cacheable preprocessing runs once per video and extracts reusable evidence before question-specific composition.
CLIP encodes 600 frames, CLAP scores 300 two-second audio windows, OCR processes 300 sampled frames (${\sim}$2.5k crops), and Whisper transcribes the full 600\,s audio track.
The half-rate OCR schedule is an algorithmic design choice: on-screen text such as road signs, digital displays, and text printed on objects is typically less dynamic than objects, events, and visual concepts, so dense 1\,FPS OCR provides limited additional evidence relative to its cost.
Under the 8-GPU schedule, the preprocessing critical path is 2.74\,s, governed by the slowest parallel branch; the TFLOP entry in the total row reports the busiest compute branch used for the latency accounting.

\paragraph{Per-query processing.}
Each query runs the Qwen3-VL-8B parser on text only, query-conditioned YOLOv8x-worldv2 when an OVD leaf appears, lightweight CLIP-text/cosine and ASR matching, and CPU composition with PASS.
The per-query selector critical path is 1.88\,s; the first-query selector cost is 4.63\,s, given by 2.74\,s cacheable preprocessing plus 1.88\,s per-query processing up to rounding.
Most video evidence is therefore amortized across questions: after caching, the query-specific path is dominated by the text parser, while OVD, scoring, and composition remain sub-second.

\paragraph{Interpretation.}
The breakdown highlights where HiMu spends computation under a fixed RTX 6000 Pro measurement protocol.
Expensive video-level evidence extraction is cacheable and parallelized across the 8 GPUs, so subsequent questions reuse CLIP, OCR, ASR, and CLAP evidence.
The uncached path for a new question is primarily the text-only tree parser, followed by lightweight query-conditioned detection, matching, and composition.

\section{Interpretability of Frame Selection}
\label{sec:interpretability}

\begin{figure*}[htbp]
  \centering
  \vspace{1em}
  \includegraphics[width=\linewidth]{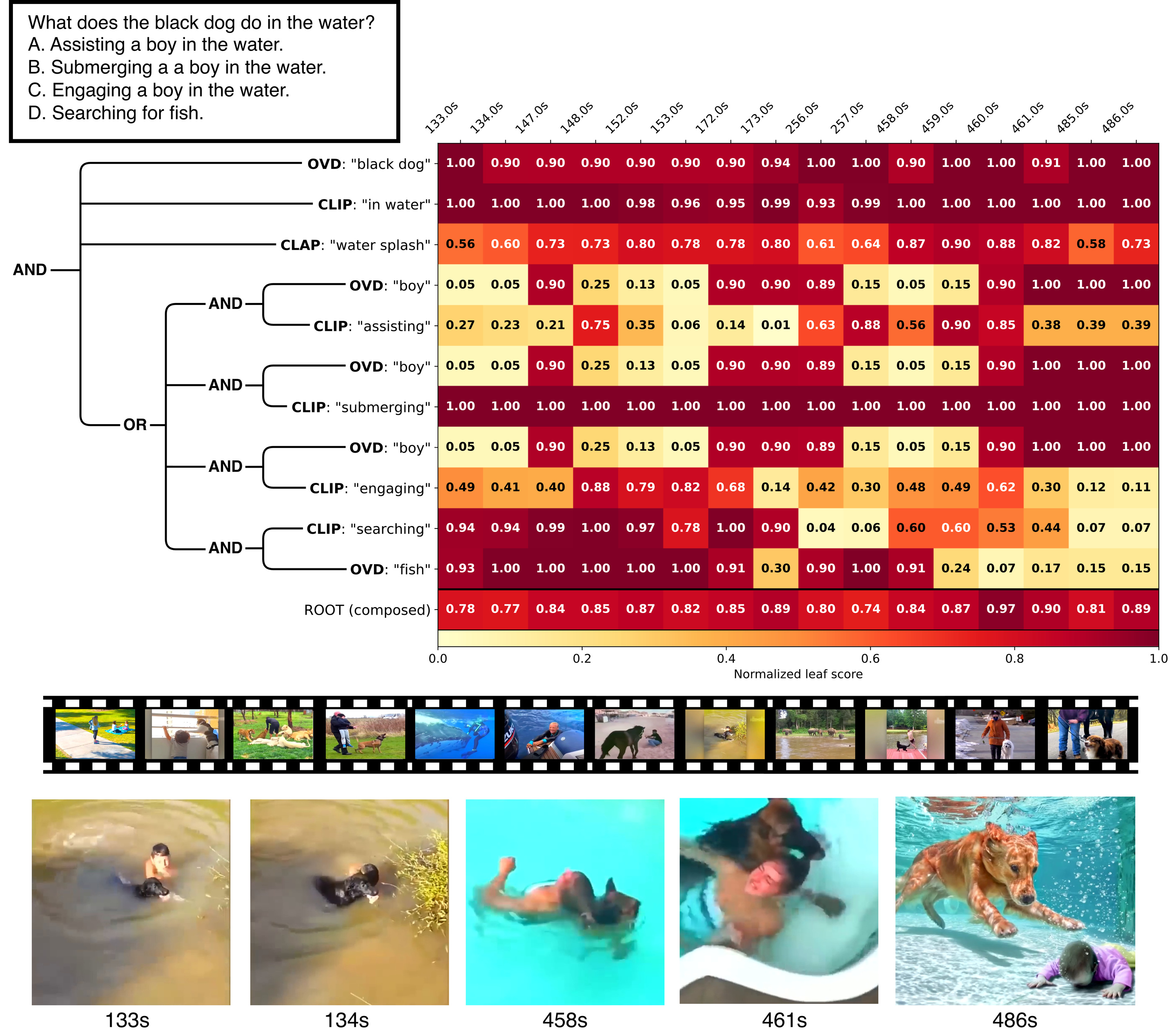}
  \caption{\textbf{Per-leaf activation heatmap for a sampled Video-MME question.}
  \emph{Top-left:} the MCQ question (``What does the black dog do in the water?'') with its four answer options.
  \emph{Centre:} the logic tree with per-leaf scores for each of the $K{=}16$ selected frames (columns, in temporal order).
  Cell colour encodes $s_e^{(i)}(t) \in [0,1]$ (dark $\approx 1$: predicate strongly satisfied; light/pale $\approx 0$: not satisfied).
  \emph{Bottom:} a filmstrip of a few of the 1\,fps sampled frames and enlarged thumbnails of five representatives from the selected frames.
  This leaf-level transparency is a structural consequence of \revised{composing named expert predicates through an explicit logic tree, and is} unavailable in similarity-based or other controlled frame selection methods.}
  \label{fig:interpretability}
\end{figure*}

A key structural benefit of \revised{composing named expert predicates through an explicit logic tree} is that every frame-selection decision is fully auditable.
Because the satisfaction score $T(t)$ is computed by a deterministic fuzzy-logic tree over named leaf scores, each frame $t$ carries an explicit attribution vector $\mathbf{a}(t) = (s_{e_1}^{(1)}(t), \ldots, s_{e_L}^{(L)}(t)) \in [0,1]^L$, where each entry records how strongly a specific expert-predicate pair fired at that frame.
This stands in sharp contrast to other controlled frame selectors: similarity-based methods (BOLT~\cite{bolt2025}, AKS~\cite{aks2025}, MDP$^3$~\cite{mdp3_2025}) collapse the entire query into a single embedding and return one opaque similarity scalar per frame, leaving no record of which part of the query drove selection or which modality was activated; structured selectors (T$^\ast$~\cite{tstar2025}, VSLS~\cite{vsls2025}) expose detection logs but are restricted to flat object queries or a fixed vocabulary of relations, so their explanations cannot reflect nested temporal-logic structure.
Beyond the controlled-selector class, multi-call agentic methods (VideoAgent~\cite{fan2024videoagent}, SeViLA~\cite{sevila2023}) produce readable reasoning traces but make their frame-selection decisions implicitly through opaque LLM outputs rather than an explicit scored ranking, so one cannot directly determine which predicate supported a selected frame or whether a failure arose from perception, temporal composition, or query decomposition.
In HiMu, by contrast, any selected frame can be immediately explained from its leaf activations. For example, ``this frame was chosen because the OVD leaf \emph{`man in suit'} scored 0.91 and the CLAP leaf \emph{`applause'} scored 0.83''; any rejected frame can be confirmed to have scored uniformly low across all predicates.

Figure~\ref{fig:interpretability} illustrates this for a representative Video-MME question (``What does the black dog do in the water?''): each column is one of the $K{=}16$ selected frames; each row is a named leaf, coloured by its normalised score $s_e^{(i)}(t) \in [0,1]$.
The shared-context leaves under the root AND node (OVD: ``black dog'' and CLIP: ``in water'') are uniformly dark across nearly all frames, confirming that the selected frames consistently depict the relevant scene.
Within the OR over the four MCQ options, option-specific leaves exhibit clearly distinct activation patterns: for instance, CLIP: ``submerging'' fires strongly (1.00) on every frame, whereas CLIP: ``searching'' and OVD: ``fish'' are dark only in the earlier temporal cluster (about 133 to 173\,s) and fade to pale in the later cluster (about 459 to 586\,s), revealing which moments support which answer candidate.
The bottom of the figure shows the corresponding frame thumbnails, letting a user visually verify the heatmap activations against the actual video content.

This interpretability has direct practical value: when HiMu answers a question incorrectly, the heatmap pinpoints whether the failure stems from a specific expert (e.g., OCR missing on-screen text), the tree structure (e.g., a SEQ operator with the wrong temporal ordering), or the query decomposition (e.g., an overly generic predicate).
Such fine-grained, per-predicate transparency is, to our knowledge, unique among current frame-selection methods and turns HiMu into a diagnostic tool rather than merely a preprocessing step, enabling practitioners to iteratively refine expert configurations, tree templates, and smoothing parameters with targeted, evidence-based feedback.